\documentclass[11pt]{article}

\usepackage{acl}

\usepackage{times}
\usepackage{latexsym}

\usepackage[T1]{fontenc}

\usepackage[utf8]{inputenc}

\usepackage{microtype}

\usepackage{inconsolata}

\usepackage{graphicx}

\title{Learning to Explain: Supervised Token Attribution from Transformer Attention Patterns}

\author{\textbf{George Mihaila} \\
  University of North Texas \\
  \texttt{\small georgemihaila@my.unt.edu} \\
  }

\begin{document}
\maketitle
\thispagestyle{plain}
\begin{abstract}
Explainable AI (XAI) has become critical as transformer-based models are deployed in high-stakes applications including healthcare, legal systems, and financial services, where opacity hinders trust and accountability. Transformers' self-attention mechanisms have proven valuable for model interpretability, with attention weights successfully used to understand model focus and behavior \cite{Xu2015ShowAA}; \cite{Wiegreffe2019AttentionIN}. However, existing attention-based explanation methods rely on manually defined aggregation strategies and fixed attribution rules \cite{Abnar2020QuantifyingAF}; \cite{Chefer2021GenericAE}, while model-agnostic approaches (LIME, SHAP) treat the model as a black box and incur significant computational costs through input perturbation. We introduce Explanation Network (ExpNet), a lightweight neural network that learns an explicit mapping from transformer attention patterns to token-level importance scores. Unlike prior methods, ExpNet discovers optimal attention feature combinations automatically rather than relying on predetermined rules. We evaluate ExpNet in a challenging cross-task setting and benchmark it against a broad spectrum of model-agnostic methods and attention-based techniques spanning four methodological families. 
\end{abstract}

\section{Introduction}
Transformer architectures \cite{Vaswani2017AttentionIA} have revolutionized Natural Language Processing, achieving state-of-the-art performance across diverse tasks. Despite their success, transformers remain complex and difficult to interpret, as predictions emerge from multiple layers of self-attention and nonlinear transformations. This opacity raises critical concerns for transparency and trustworthiness, particularly when models are deployed in sensitive domains such as healthcare diagnostics, legal decision support, and financial risk assessment \cite{DoshiVelez2017TowardsAR}. Rigorous, human-centered interpretability has therefore become essential for accountable deployment of these powerful yet opaque models.
Existing XAI approaches fall into two main categories. Model-agnostic methods such as LIME \cite{Ribeiro2016WhySI} and SHAP \cite{Lundberg2017AUA} treat models as black boxes, offering broad applicability but ignoring internal reasoning and incurring significant computational costs. In NLP specifically, these perturbation-based methods face a fundamental challenge: masking or replacing words creates semantically invalid examples that lose linguistic meaning \cite{Feng2018PathologiesON}.
The transformer's self-attention mechanism offers an alternative explanation source. Attention weights have proven valuable for interpretability since their introduction \cite{Bahdanau2014NeuralMT}; \cite{Xu2015ShowAA}; \cite{Wiegreffe2019AttentionIN}, leading to numerous attention-based explanation methods \cite{Abnar2020QuantifyingAF};  \cite{Qiang2022AttCATET}. However, these methods rely on hand-crafted heuristic rules or fixed aggregation functions to extract explanations from attention patterns.


Our contribution treats attention-based explanation as a learning problem. This learned approach generalizes across tasks and consistently outperforms established baseline methods while maintaining computational efficiency.
Our work contributes: (1) a novel framework that learns attention-based explanations from human rationales rather than relying on fixed heuristics, (2) demonstration of cross-task explainability generalization in NLP, (3) comprehensive ablation studies justifying architectural choices, and (4) extensive empirical evaluation showing state-of-the-art performance.

\section{Related Work}

\subsection{Model-Agnostic Explanation Methods}

Model-agnostic techniques treat models as black boxes, requiring only input-output access. LIME \cite{Ribeiro2016WhySI} learns local linear approximations by probing models on perturbed inputs, while SHAP \cite{Lundberg2017AUA} provides game-theoretic Shapley value-based attributions with formal guarantees of consistency and local accuracy.
While offering broad applicability, these methods face limitations in NLP contexts. They ignore model internals including transformer attention mechanisms, incur significant computational costs through extensive model queries, and perturbation-based approaches can produce out-of-distribution inputs that confound causal interpretation \cite{Feng2018PathologiesON}.

\subsection{Attention as Explanation}
Attention mechanisms have proven valuable for interpretability since their introduction, with attention weights widely visualized to understand model focus \cite{Bahdanau2014NeuralMT, Xu2015ShowAA, Vig2019AMV}. Research has shown that attention patterns capture meaningful linguistic phenomena including syntactic dependencies and semantic relationships \cite{Wiegreffe2019AttentionIN}. This motivates attention-based explanation methods that leverage these internal signals, which are common to transformer architectures and unavailable to model-agnostic approaches.

\subsection{Gradient-Based and Attention-Based Methods}

Gradient-based methods leverage backpropagation to attribute importance. Integrated Gradients \cite{IntegratedGradients} accumulates gradients along paths from baseline to actual inputs, satisfying axioms including sensitivity and implementation invariance, but operates independently of attention mechanisms.

Methods combining gradients with attention leverage complementary signals from  backpropagation and attention mechanisms. Gradient-weighted approaches multiply attention scores by loss gradients, analogous to Grad-CAM in computer vision \cite{Selvaraju2016GradCAMVE, Barkan2021GradSAMET}. AttCAT \cite{Qiang2022AttCATET} integrates attention weights, token gradients, and skip connections. Layer-wise Relevance Propagation (LRP) methods conservatively redistribute output scores through systematic backpropagation \cite{Bach2015OnPE}. Transformer-LRP \cite{Transformer‑LRP} adapted LRP for transformers, Generic Attention Explainability (GAE) \cite{Chefer2021GenericAE} extended it to encoder-decoder models, and Mask-LRP \cite{Mask-LRP} refined the approach by masking uninformative heads.

Attention aggregation approaches combine attention signals from multiple heads or layers using predetermined rules. Attention Rollout and Attention Flow \cite{attrollout} aggregate attention across layers to trace token influence, addressing entangled representations in deeper layers \cite{Brunner2019OnII}. These approaches improved correlation with ablation studies through multi-layer composition rules.

These methods extract explanations through predetermined procedures—gradient computations, propagation algorithms, or aggregation functions—that are specified apriori based on mathematical or heuristic principles. Each method embodies specific assumptions about how to derive token importance from model internals.

\subsection{Explanation Evaluation}

Evaluation of explanation quality faces fundamental challenges. Faithfulness metrics assess whether explanations reflect model reasoning through perturbation \cite{DeYoung2019ERASERAB}, but suffer from critical problems in NLP: token removal creates out-of-distribution inputs \cite{Feng2018PathologiesON}, architectural redundancy enables compensation where alternative pathways maintain predictions despite removing important features \cite{Voita2019AnalyzingMS, Serrano2019IsAI}, and circular validation without ground truth undermines validity \cite{Hooker2018ABF}. Plausibility metrics, which measure alignment with human rationales \cite{Jacovi2020TowardsFI}, provide a more principled evaluation framework for high-stakes applications where human interpretability is paramount \cite{DoshiVelez2017TowardsAR}. Our approach embraces this paradigm by learning explanations directly from human rationales.

\subsection{Positioning ExpNet}

ExpNet builds upon attention-based explanation methods, leveraging the insight that transformer attention patterns contain valuable signals for interpretability \cite{Bahdanau2014NeuralMT, Wiegreffe2019AttentionIN, attrollout}. Rather than applying fixed aggregation rules or heuristics, ExpNet treats explanation extraction as a supervised learning problem, training on human-annotated rationales to discover which attention configurations correlate with human-judged token importance. This approach enables cross-task generalization and direct optimization for human-aligned explanations while reducing the need for human rationales when applied to new datasets.

\section{Methodology}

\subsection{Overview and Base Model}

We use BERT-base \cite{Devlin2019BERTPO} as our transformer model for all experiments. 
BERT processes input text by tokenizing it into subword units and prepending a special 
classification token \texttt{[CLS]} to each sequence. During classification, BERT's 
final hidden state corresponding to \texttt{[CLS]} is passed to a classification head 
to produce the prediction. This design makes \texttt{[CLS]} a natural aggregation point: 
through self-attention across all layers, \texttt{[CLS]} gathers task-relevant 
information from input tokens.

Our goal is to learn which input tokens are important for BERT's decisions by analyzing 
attention patterns involving \texttt{[CLS]}. Specifically, we train a lightweight neural 
network—ExpNet—that maps BERT's self-attention patterns to token-level importance scores 
aligned with human rationales. ExpNet treats BERT as a frozen feature extractor: we do 
not modify BERT's weights, only extract and process its attention patterns. While we use BERT-base for experimental validation, ExpNet's approach generalizes to  other encoder-based transformers with similar aggregation mechanisms (e.g., RoBERTa,  ALBERT, DeBERTa) and can be adapted to decoder-based or encoder-decoder-base models with appropriate  modifications.

\subsection{Attention-Based Feature Extraction}

We extract features from BERT's final self-attention layer (layer 12), which contains 
$H=12$ attention heads. For each token $j$ in the input sequence, we extract two types 
of attention-based features from each head $h \in \{1, \ldots, H\}$:

\paragraph{Task-to-Token Attention:} 
The attention weight from \texttt{[CLS]} 
to token $j$ 
, denoted 
$A^{(h)}_{\texttt{[CLS]}, j}$. This measures how strongly BERT's aggregation mechanism 
attends to token $j$ when forming its prediction. High values suggest \texttt{[CLS]} 
draws information from token $j$.

\paragraph{Token-to-Task Attention:} 
The attention weight from token $j$ 
to \texttt{[CLS]} 
, denoted 
$A^{(h)}_{j,\texttt{[CLS]}}$. This measures how much token $j$ attends to the task 
representation. High values suggest token $j$ is contextualizing itself relative to the 
classification objective.

These two perspectives capture the non-symmetric nature of self-attention relationships, 
where $A_{i,j} \neq A_{j,i}$. A token may be important either because \texttt{[CLS]} 
focuses on it (task-to-token: \texttt{[CLS]} extracts information from the token) or because the 
token focuses on \texttt{[CLS]} (token-to-task: the token contextualizes itself relative 
to the task). Prior attention-based explanation methods 
often averaged attention weights across heads \cite{Wiegreffe2019AttentionIN} or used 
only one attention direction, potentially discarding information valuable for cross-task 
generalization. By preserving individual head contributions from both directions, ExpNet 
learns which attention patterns indicate importance—patterns that transfer across tasks 
because attention heads capture general linguistic relationships (syntax, semantics, 
discourse structure) independent of specific classification objectives 
\cite{Voita2019AnalyzingMS, Clark2019WhatDB}.

These two perspectives capture asymmetric attention relationships: a token may be 
important either because \texttt{[CLS]} focuses on it (task-to-token) or because it 
focuses on \texttt{[CLS]} (token-to-task). Prior attention-based explanation methods 
often averaged attention weights across heads \cite{Wiegreffe2019AttentionIN} or used 
only one attention direction. By preserving individual head contributions from both 
directions, we enable ExpNet to learn which attention patterns are most indicative of 
token importance.

\begin{figure*}[t]
\centering
\includegraphics[width=0.7\textwidth, height=0.9\textheight, keepaspectratio]{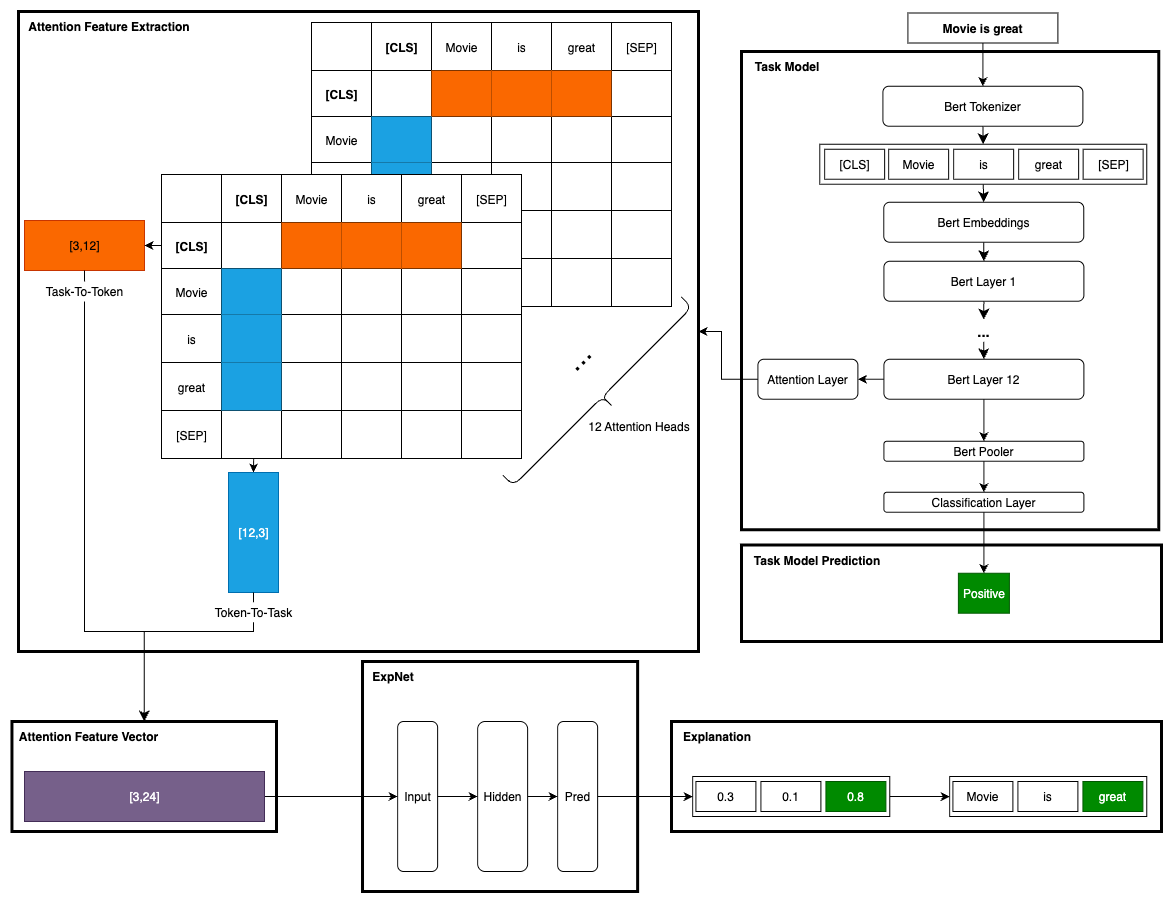}
\caption{ExpNet complete pipeline from BERT's attention patterns to importance predictions.}
\label{fig:expnet_architecture}
\end{figure*}

Each token $j$ is represented by a feature vector concatenating all attention values:
\begin{equation}
\begin{aligned}
f_j = \bigl[ &A_{\texttt{[CLS]},j}^{(1)},\,\ldots,\,A_{\texttt{[CLS]},j}^{(H)}, \\
             &A_{j,\texttt{[CLS]}}^{(1)},\,\ldots,\,A_{j,\texttt{[CLS]}}^{(H)} \bigr]
\end{aligned}
\label{eq:token_features}
\end{equation}
For BERT-base with $H=12$ heads, $f_j$ is a 24-dimensional vector. The entire ExpNet pipeline is illustrated in Figure \ref{fig:expnet_architecture}.

\subsection{ExpNet Architecture}

ExpNet is a simple feed-forward network that maps the feature vector $f_j$ to an 
importance score for token $j$. The architecture consists of:
\begin{itemize}
    \item Input layer: 24 dimensions (2$H$ attention features)
    \item Hidden layer: 16 units with ReLU activation
    \item Output layer: Single sigmoid unit producing importance probability $\hat{y}_j \in [0,1]$
\end{itemize}

Formally, ExpNet computes:
\begin{equation}
\begin{aligned}
\mathbf{h}_j &= \text{ReLU}(\mathbf{W}_1 f_j + \mathbf{b}_1) \\
\hat{y}_j &= \sigma(\mathbf{w}_2^{\top} \mathbf{h}_j + b_2)
\end{aligned}
\end{equation}
where $\sigma(x) = 1/(1+e^{-x})$ is the sigmoid function, and $\hat{y}_j$ represents 
the predicted probability that token $j$ is important.

\subsection{Training Procedure}

\paragraph{Supervision from Human Rationales:}
We train ExpNet as a token-level binary classifier using human-annotated rationales. 
For each training example, humans mark which words justify the label. Since human 
annotations are at the word level while BERT operates on subword tokens, we project 
labels: if any subtoken of a word is marked important by humans, all subtokens of that 
word receive label $y_j=1$; otherwise $y_j=0$.

\paragraph{Training on Correct Predictions Only:}
We train ExpNet exclusively on instances where BERT's prediction is correct. This 
ensures the explainer learns attention patterns associated with successful reasoning. 
Training on mispredictions could introduce noise, as attention patterns may be 
unreliable when the model makes errors \cite{DeYoung2019ERASERAB}. At inference time, 
ExpNet can still explain any prediction (correct or incorrect).

\paragraph{Focal Loss for Class Imbalance:}
Token importance labels are highly imbalanced: most tokens are not important. To 
address this, we use focal loss \cite{Lin2017FocalLF} instead of cross-entropy:
\begin{equation}
\mathcal{L}_{\text{focal}} = -\alpha_t (1 - p_t)^{\gamma} \log(p_t)
\end{equation}
where $p_t = \hat{y}_j$ if $y_j=1$, else $p_t = 1-\hat{y}_j$. The modulating factor 
$(1-p_t)^{\gamma}$ down-weights easy examples, focusing learning on hard-to-classify 
tokens. We set $\alpha=0.6$ and $\gamma=2$ following \cite{Lin2017FocalLF}.

\paragraph{Inference and Thresholding:}
At test time, we apply a threshold of 0.5 to $\hat{y}_j$ to produce binary importance 
predictions. To ensure ExpNet identifies at least one important token per example (even 
when all scores fall below 0.5), we select the highest-scoring token as important if no 
tokens exceed the threshold. This reflects the assumption that every prediction has at 
least some supporting evidence and prevents degenerate cases with empty explanations.

\paragraph{Hyperparameters:}
We train ExpNet for 50 epochs using Adam optimizer with learning rate 0.001 and batch 
size 32. No learning rate schedule or warmup is used. These hyperparameters were chosen 
to ensure stable convergence across all three datasets.

\section{Experiments}

We evaluate ExpNet on three NLP classification tasks (Table \ref{tab:datasets_overview}), examining the challenging scenario of cross-domain (cross-task) explainability. This chapter describes the datasets, experimental training setups, baseline explanation methods, and evaluation metrics.

\subsection{Datasets and Annotation}

We evaluate ExpNet on three NLP classification tasks with distinct characteristics. For each task, we fine-tune a separate BERT-base classifier \cite{Devlin2019BERTPO} that serves as the frozen feature extractor.


\begin{table*}
  \centering
  \begin{tabular}{llll}
    \hline
    \textbf{Dataset} & \textbf{Task} & \textbf{Source} & \textbf{Rationales} \\
    \hline
    HateXplain & Hate speech & Social media & Human \\
    SST-2 & Sentiment & Movie reviews & LLM + Human \\
    CoLA & Grammaticality & Linguistic & LLM + Human \\
    \hline
  \end{tabular}
  \caption{\label{tab:datasets_overview}
    Dataset overview and annotation sources.
  }
\end{table*}

\textbf{HateXplain} \cite{Mathew2020HateXplainAB} includes native human rationales. We use only hate and offensive examples. \textbf{SST-2} \cite{Socher2013RecursiveDM} and \textbf{CoLA} \cite{warstadt2018neural} lack word-level rationales. We generated training annotations using Claude-3.5-Sonnet following a systematic process: calibration on HateXplain, prompt refinement with human-annotated 30-example seed sets, and large-scale annotation. Critically, test sets for both SST-2 and CoLA were independently annotated by two human annotators, achieving substantial inter-annotator agreement (Table \ref{tab:dataset_splits}). Complete annotation methodology and reliability analysis are provided in Appendix~\ref{app:annotation}.


\begin{table*}
  \centering
  \begin{tabular}{lccl}
    \hline
    \textbf{Dataset} & \textbf{Train/Val} & \textbf{Test} & \textbf{Test Annotation} \\
    \hline
    SST-2       & 5,971/395 & 436 & Human (2 ann., $\alpha$=0.84) \\
    CoLA        & 2,426/101 & 161 & Human (2 ann., $\alpha$=0.64) \\
    HateXplain  & 8,042/734 & 752 & Human (3 ann., $\alpha$=0.46) \\
    \hline
  \end{tabular}
  \caption{\label{tab:dataset_splits}
    Final dataset splits. CoLA uses only unacceptable examples; HateXplain excludes neutral class.
  }
\end{table*}

Following standard practice \cite{DeYoung2019ERASERAB}, we apply \textit{correct-prediction filtering}: explanations are evaluated only on instances where the classifier's prediction matches the gold label, ensuring we explain true decision-making rather than error patterns.

\subsection{Cross-Task Generalization Experimental Design}

The evaluation employs a \textit{leave-one-task-out} protocol where ExpNet is trained on the combined rationale data from two tasks and evaluated on the held-out third task. No task-specific hyperparameter tuning is performed; all training parameters remain fixed across experiments to ensure fair cross-task generalization assessment. The three experimental configurations are:

\begin{itemize}
    \item Train on (SST-2 + CoLA) $\rightarrow$ Test on HateXplain
    \item Train on (SST-2 + HateXplain) $\rightarrow$ Test on CoLA
    \item Train on (CoLA + HateXplain) $\rightarrow$ Test on SST-2
\end{itemize}

This cross-task design demonstrates that ExpNet learns generalizable explanation patterns rather than task-specific heuristics.

\subsection{Baseline Explanation Methods}
\label{sec:baselines}

We benchmark ExpNet against thirteen established explainability methods representing fundamentally different approaches to generating explanations. \textbf{LIME} \cite{Ribeiro2016WhySI} learns a sparse linear surrogate model around each instance by probing the classifier on perturbed inputs, treating the model as a black box. Similarly, \textbf{SHAP} \cite{Lundberg2017AUA} adopts a model-agnostic approach, framing attribution as a Shapley-value problem from cooperative game theory to estimate each feature's marginal contribution. Unlike these black-box methods, \textbf{Integrated Gradients} \cite{Sundararajan2017AxiomaticAF} leverages model internals by computing gradients along a straight-line path from a baseline to the input, integrating to obtain attributions. Building on this gradient-based foundation, \textbf{Layer-wise Relevance Propagation (LRP)} \cite{Bach2015OnPE} back-propagates output relevance to the input layer using conservation rules, while \textbf{FullLRP} \cite{Voita2019AnalyzingMS} modifies this by stopping propagation at the final transformer block rather than continuing to raw embeddings, making it more suitable for transformer architectures. \textbf{CAM} \cite{Selvaraju2016GradCAMVE} takes a different gradient-based approach by leveraging the final hidden layer to localize important tokens, and \textbf{Grad-CAM} \cite{Barkan2021GradSAMET} extends this idea by using gradients with respect to the final hidden layer to weight token contributions.

Several methods combine multiple signals for more robust attribution. \textbf{Generic Attention Explainability (GAE)} \cite{Chefer2021GenericAE} merges attention gradients with relevance propagation rules to obtain token attributions, effectively bridging gradient-based and propagation-based approaches. \textbf{Mask-LRP (MGAE)} \cite{Mask-LRP} extends the LRP framework by first identifying and masking non-informative attention heads (e.g., syntactic or positional heads), then propagating relevance only through retained heads for cleaner attributions. \textbf{AttCAT} \cite{Qiang2022AttCATET} fuses per-block attention weights with input-token gradients to form attentive class-activation maps, combining attention mechanisms with gradient-based localization. Finally, two methods rely purely on attention mechanisms without gradients or propagation: \textbf{RawAtt} \cite{attrollout} directly uses averaged self-attention weights from the last layer as token-importance scores, representing the simplest attention-based approach, while \textbf{Attention Rollout} \cite{attrollout} builds on this by aggregating attention weights across all layers through matrix multiplication to capture long-range dependencies throughout the network. Finally, we include a naive \textbf{RandomBaseline}, which ignores model internals and independently marks each token as important with probability equal to the empirical positive rate in the training data, providing a trivial lower bound for comparison.

To enable fair comparison with binary human rationales, we binarize continuous baseline scores using top-K selection following standard practice \cite{DeYoung2019ERASERAB}, where K is the average rationale length per dataset (K=3 for SST-2, K=15 for CoLA, K=9 for HateXplain). Detailed binarization procedures are provided in Appendix~\ref{app:baselines}.

\section{Results}

We assess explanation quality using token-level F1-score and AUROC, measuring alignment between predicted rationales and human-annotated ground truth. \textbf{F1-score} is appropriate for this task as it balances precision and recall when identifying important tokens, penalizing both missed rationales and spurious highlights. \textbf{AUROC} provides a threshold-independent assessment of ranking quality, evaluating how consistently methods order true rationale tokens ahead of non-rationale tokens across all decision boundaries. Together, these metrics capture both binary classification accuracy (F1) and continuous ranking performance (AUROC) of explanation methods. We report dataset-level metrics following standard practice in rationale evaluation \cite{DeYoung2019ERASERAB}. Complete metric definitions and formulas are provided in Appendix~\ref{app:metrics}.

\subsection{Quantitative Evaluation}

We compare ExpNet against thirteen established baseline methods. This comprehensive comparison ensures robust evaluation against both classic post-hoc explainers and recent state-of-the-art methods.

\begin{table*}[t]
  \centering
  \begin{tabular}{llll}
    \hline
    \textbf{Explainer}      & \textbf{SST-2}          & \textbf{CoLA}           & \textbf{HateXplain} \\
    \hline
    RandomBaseline          & $0.258$                 & $0.387$                 & $0.293$           \\
    SHAP                    & $0.330 \pm 0.033$       & $0.330 \pm 0.056$       & $0.276 \pm 0.007$ \\
    LIME                    & $0.347 \pm 0.033$       & $0.323 \pm 0.053$       & $0.290 \pm 0.007$ \\
    Integrated Gradient     & $0.287 \pm 0.032$       & $0.342 \pm 0.054$       & $0.345 \pm 0.008$ \\
    RawAt                   & $0.327 \pm 0.029$       & $0.353 \pm 0.054$       & $0.362 \pm 0.007$ \\
    Rollout                 & $0.133 \pm 0.025$       & $0.347 \pm 0.055$       & $0.356 \pm 0.007$ \\
    LRP                     & $0.339 \pm 0.030$       & $0.355 \pm 0.054$       & $0.372 \pm 0.008$ \\
    FullLRP                 & $0.218 \pm 0.030$       & $0.336 \pm 0.055$       & $0.346 \pm 0.007$ \\
    GAE                     & $0.350 \pm 0.030$       & $0.354 \pm 0.053$       & $0.391 \pm 0.007$ \\
    CAM                     & $0.243 \pm 0.032$       & $0.355 \pm 0.054$       & $0.332 \pm 0.007$ \\
    GradCAM                 & $0.234 \pm 0.031$       & $0.356 \pm 0.056$       & $0.396 \pm 0.008$ \\
    AttCAT                  & $0.280 \pm 0.034$       & $0.345 \pm 0.055$       & $0.340 \pm 0.007$ \\
    MGAE                    & $0.350 \pm 0.030$       & $0.354 \pm 0.053$       & $0.391 \pm 0.007$ \\
    \hline
    ExpNet                  & $\mathbf{0.398 \pm 0.024}$ & $\mathbf{0.468 \pm 0.079}$ & $\mathbf{0.473 \pm 0.007}$ \\
    \hline
  \end{tabular}
  \caption{\label{quantitative-results}
    F1-score results for ExpNet compared to baseline methods under cross-dataset evaluation. ExpNet is trained on two datasets and evaluated on the held-out third dataset: SST-2 performance when trained on CoLA + HateXplain, CoLA performance when trained on SST-2 + HateXplain, and HateXplain performance when trained on SST-2 + CoLA. For all trained methods we report F1-Score with ± 95\% confidence intervals; RandomBaseline is reported as a single deterministic run.
  }
\end{table*}

Table~\ref{quantitative-results} reports token-level F1 scores with confidence intervals across the three benchmarks under the cross-dataset setting. ExpNet achieves the highest F1 across all three tasks, demonstrating substantially better alignment with human rationales than all thirteen baselines. The cross-task generalization is particularly notable: ExpNet was never trained on the held-out task, yet consistently outperforms methods specifically tuned for those datasets. On CoLA, ExpNet achieves F1 of $0.468 \pm 0.079$, representing a 31\% relative improvement over the best baseline (GradCAM: 0.356). On HateXplain, ExpNet reaches $0.473 \pm 0.007$, outperforming the strongest baseline (GradCAM: 0.396) by 19\%. Even on SST-2, where baseline performance is more competitive, ExpNet's F1 of $0.398 \pm 0.024$ exceeds the best baseline (GAE/MGAE: 0.350) by 14\%. Detailed results are provided in Appendix \ref{sec:performance_metrics}.

The baseline results reveal distinct performance patterns across explainer categories. On HateXplain, gradient-based methods (GradCAM: $0.396 \pm 0.008$) and propagation-based techniques (GAE/MGAE: $0.391 \pm 0.007$) lead among baselines. Model-agnostic methods show more variable performance, with LIME achieving $0.347 \pm 0.033$ on SST-2 but only $0.290 \pm 0.007$ on HateXplain, likely due to sampling noise and difficulty capturing complex token interactions in subjective hate speech contexts. Interestingly, on SST-2, Integrated Gradient ($0.287 \pm 0.032$) underperforms even raw attention weights (RawAt: $0.327 \pm 0.029$), suggesting that simple attention patterns can sometimes be more indicative than gradient-based approximations for certain tasks.

ExpNet's superior performance stems from direct supervision on human rationales. Methods like RawAt and Rollout \cite{Jain2019AttentionIN} rely on hand-crafted aggregation heuristics that demonstrate can be misleading. Gradient-propagation approaches such as MGAE and GAE follow fixed relevance rules, but without training on human rationales they miss subtle language patterns that humans recognize as salient. Even sophisticated hybrids like AttCAT, which combines attention weights, skip connections, and feature gradients, fall short of ExpNet's learned mapping from attention to importance.

\begin{figure*}[t]
  \includegraphics[width=0.32\linewidth]{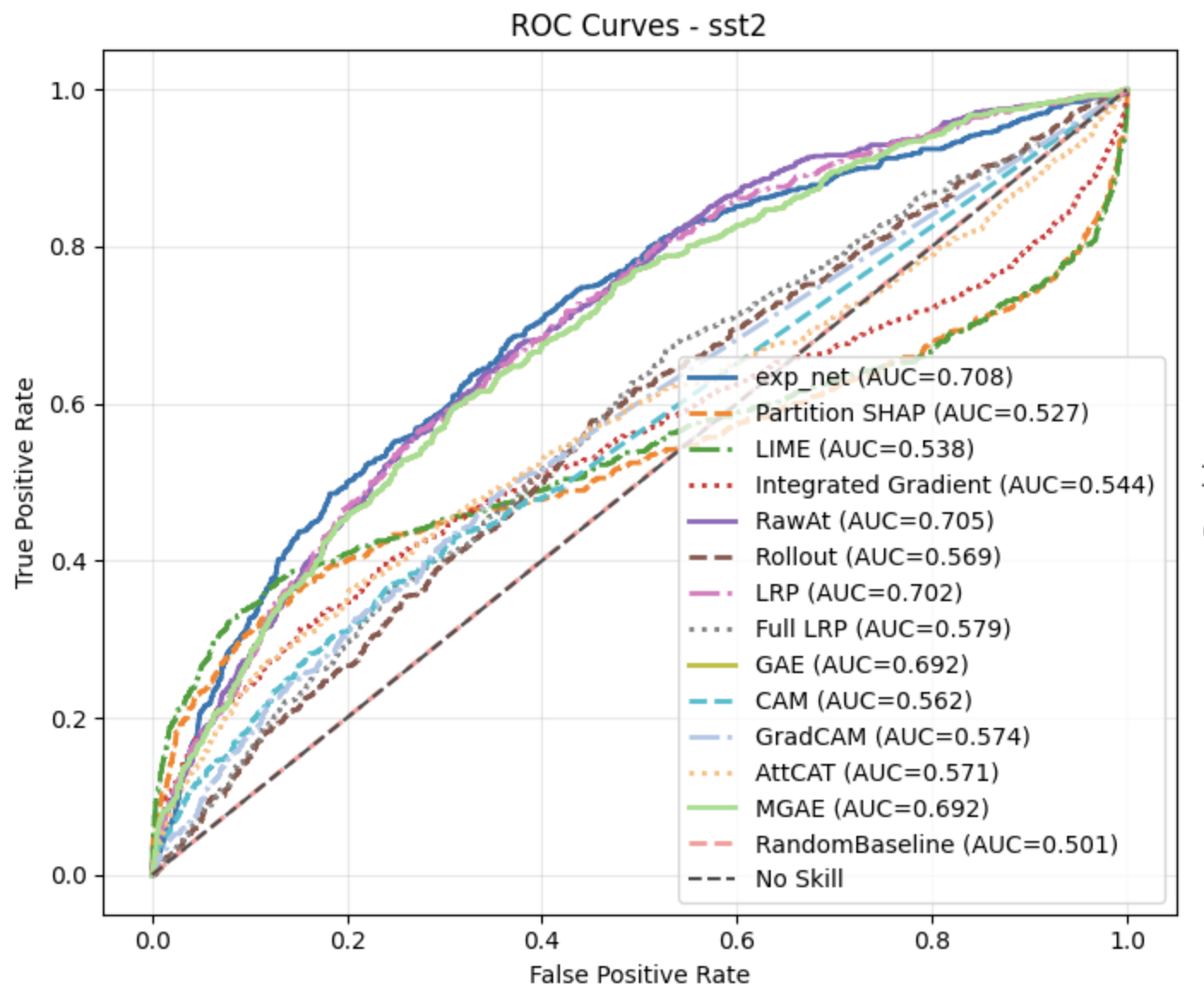} \hfill
  \includegraphics[width=0.32\linewidth]{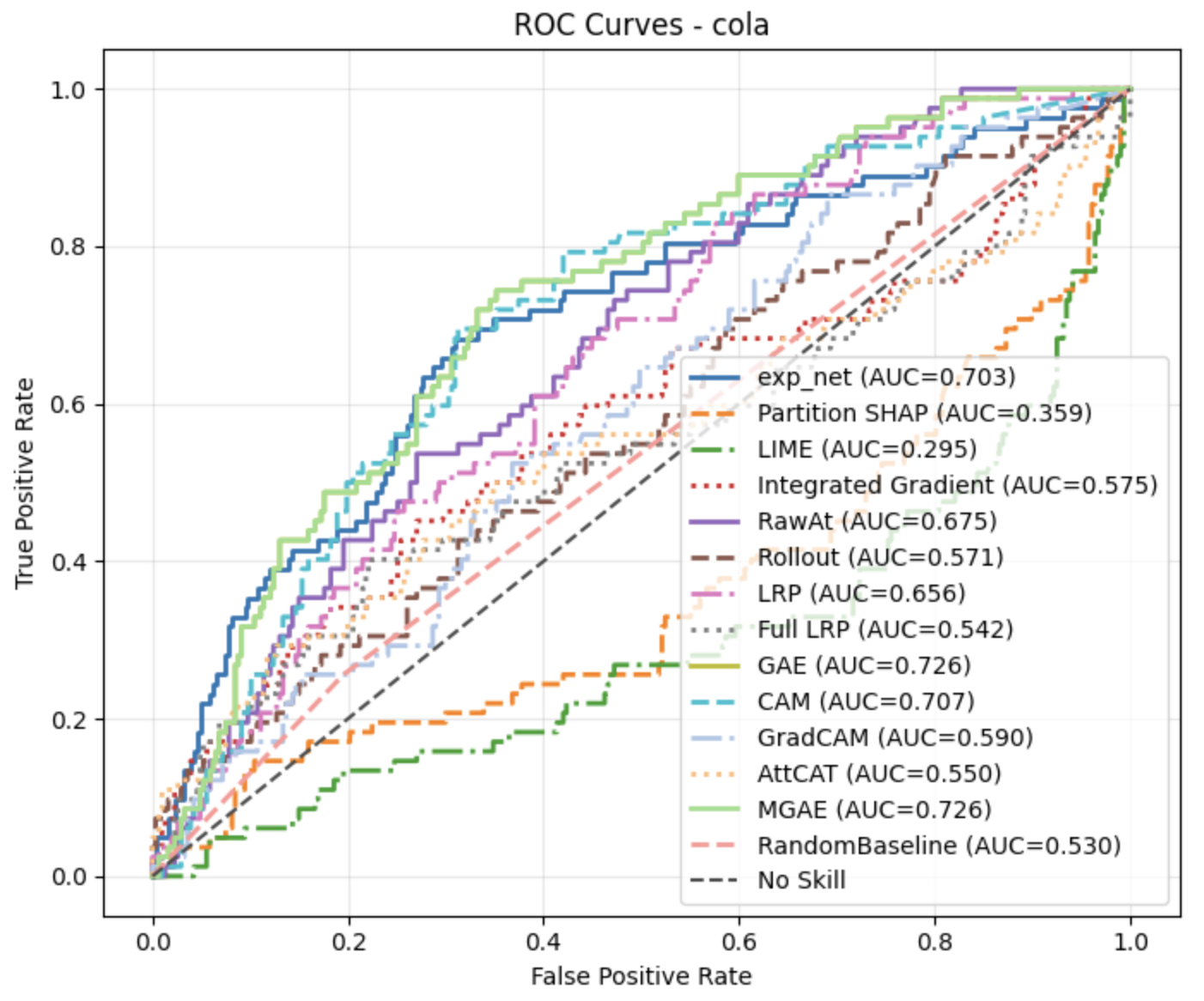} \hfill
  \includegraphics[width=0.32\linewidth]{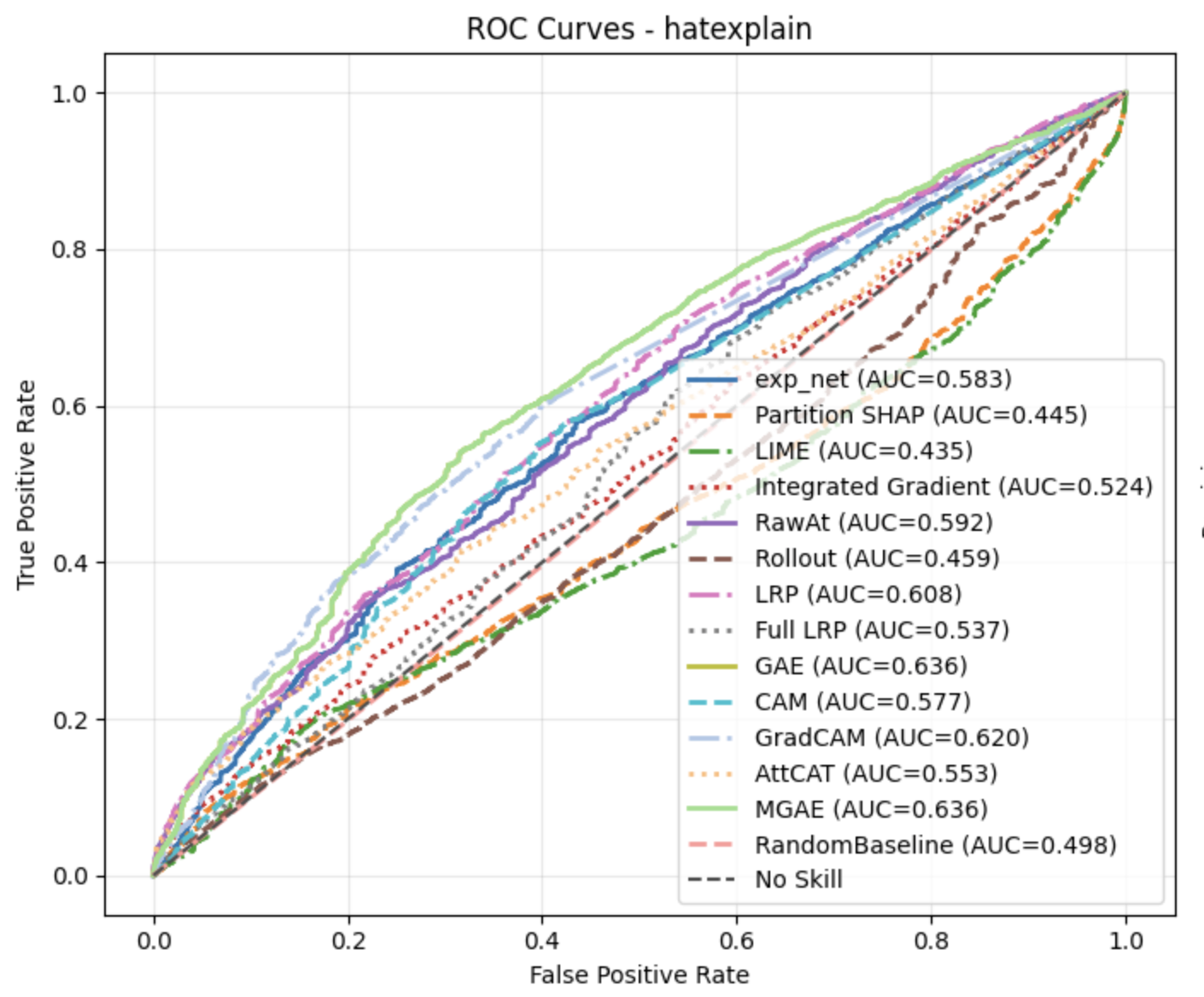}
  \caption{\label{auroc}
  AUROC values across datasets show ExpNet consistently achieves competitive ranking performance (often above 0.7), generally assigning higher scores to important tokens and lower scores to unimportant ones more effectively than most baselines.}
\end{figure*}

Figure~\ref{auroc} shows the AUROC of each explainer's token-level ranking on held-out datasets. While ExpNet performs competitively, achieving AUROC values consistently above 0.7, it does not always achieve the top AUROC. ExpNet is the top performer on SST-2, but some gradient-based baselines achieve marginally higher AUROC on CoLA and HateXplain. This pattern reveals an important distinction: ExpNet's threshold-based F1 scores are consistently the highest, but its continuous importance scores are not always the most separable in a ranking sense. ExpNet excels at selecting the correct set of important tokens (high precision and recall for the binary decision), while certain gradient-based methods may preserve relative ranking quality across all tokens more effectively. This complementary strength reflects ExpNet's training objective, which optimizes for accurate binary rationale prediction rather than perfect ranking across all importance levels.

Despite this AUROC trade-off, ExpNet remains highly robust across all three tasks, outperforming attention-only baselines by substantial margins and ranking competitively against more complex attribution methods like AttCAT and GAE. The consistency of ExpNet's F1 improvements across diverse tasks—sentiment analysis (SST-2), grammatical acceptability (CoLA), and hate speech detection (HateXplain)—demonstrates that the learned attention-to-rationale mapping generalizes beyond dataset-specific artifacts to capture fundamental patterns in how human-important features manifest in transformer attention.

\subsection{Ablation Study}
\label{sec:ablation}

To understand the contribution of different architectural components in ExpNet, we conduct ablation studies across all three datasets using various feature combinations. Table~\ref{tab:ablation} presents F1-score results comparing five variants: Hidden State (HS) features alone, combined HS and ExpNet's attention features (HS + ExpNet), task-to-token attention only, token-to-task attention only, and the complete ExpNet architecture.


\begin{table*}
  \centering
  \begin{tabular}{lccc}
    \hline
    \textbf{Method} & \textbf{SST-2} & \textbf{CoLA} & \textbf{HateXplain} \\
    \hline
    HS & 0.352 & 0.198 & 0.334 \\
    HS + ExpNet's Attention & 0.268 & 0.276 & 0.362 \\
    Task-to-Token Attention & 0.397 & 0.391 & 0.484 \\
    Token-to-Task Attention & 0.269 & 0.384 & \textbf{0.503} \\
    ExpNet (Full) & \textbf{0.398} & \textbf{0.468} & 0.473 \\
    \hline
  \end{tabular}
  \caption{\label{tab:ablation}
    Ablation study F1-scores across three datasets.
  }
\end{table*}

\textbf{Task-Specific Patterns.} On SST-2, task-to-token attention alone (0.397) nearly matches full ExpNet (0.398), indicating sentiment analysis relies primarily on task-driven attention patterns. CoLA demonstrates ExpNet's strongest advantage, with the full model (0.468) substantially outperforming the best component (0.391) showing that grammatical analysis requires the complete bidirectional mechanism. HateXplain reveals that token-to-task attention (0.503) outperforms full ExpNet (0.473), suggesting hate speech detection benefits most from bottom-up attention where offensive tokens directly influence classification.

\textbf{Cross-Dataset Insights.} While ExpNet does not achieve the best individual performance on all datasets, it provides consistent, stable performance across diverse domains. This validates the architectural design for task-agnostic generalization.

\subsection{Computational Efficiency}


Beyond explanation quality, computational efficiency is critical. ExpNet is substantially faster than most baselines while maintaining superior explanations, since it only requires a single forward pass of a lightweight MLP over pre-computed attention features, with no gradients, sampling, or perturbations at inference time (see Appendix~\ref{sec:computational_details} for throughput details).

\section{Conclusion}

ExpNet shows that transformer attention can be systematically refined into human-aligned explanations through supervised learning. Across three NLP tasks—SST-2 (sentiment), CoLA (acceptability), and HateXplain (hate speech)—ExpNet consistently outperforms all thirteen baselines, with F1 gains of 13.9\%, 24.8\%, and 19.4\% over the best method in each domain. It also achieves task-agnostic generalization: training on two datasets and evaluating on the third still yields superior performance to heuristic- or perturbation-based methods, while remaining computationally efficient enough for real-time deployment.

ExpNet exhibits complementary strengths across metrics: it achieves the highest F1 scores on all tasks, while its AUROC is strongest on SST-2 and competitive on CoLA and HateXplain. Future work can extend this framework to decoder-only transformers (e.g., GPT-style LLMs) to enable efficient, interpretable attributions in modern large language models.





\section*{Limitations}
ExpNet relies on human-annotated rationales during training, which constrains its immediate applicability to tasks with existing explanation datasets. However, our cross-dataset experiments demonstrate that ExpNet trained on available annotated tasks can generalize to new domains, suggesting a path forward: training on diverse annotated datasets enables zero-shot explanation on unannotated tasks. Future work should systematically evaluate performance across broader domain ranges to characterize generalization boundaries.

Our experiments focus on encoder-only BERT-base models and datasets ranging from 872 to 1,924 instances. While this scope enables controlled evaluation, ExpNet's behavior on larger-scale datasets, more complex encoder architectures, and decoder-only models (including modern LLMs) remains an open empirical question. Extending to these settings represents a natural next step rather than a fundamental limitation; the core approach of learning from attention patterns generalizes architecturally, though implementation details may require adaptation.

ExpNet operates at the token level, which naturally aligns with transformer tokenization but may not capture phrase-level semantics for multi-word expressions like "not bad" or "over the moon." This represents a common challenge in token-attribution methods rather than an ExpNet-specific issue. Extensions to model phrasal importance explicitly could enhance semantic alignment, though this introduces additional complexity in aggregating token-level attention signals.

The framework assumes attention mechanisms provide sufficient signal for explanation, which holds for many transformer architectures but may be less complete for models that distribute computation heavily through feed-forward layers or skip connections. This attention-centric design choice offers computational efficiency and interpretability but naturally focuses on one component of transformer reasoning. Hybrid approaches combining attention with hidden state or gradient information could broaden coverage, as our ablation studies suggest different architectural components contribute variably across tasks.

ExpNet's explanations necessarily reflect the characteristics of training rationales, including annotator subjectivity and any biases present in human judgments. Research shows that different annotators highlight different tokens \cite{Herrewijnen2024HumanannotatedRA}, particularly across demographic groups. Rather than viewing this as a flaw specific to ExpNet, we recognize it as an inherent property of learning from human demonstrations: the model learns patterns from its training signal. This means ExpNet's outputs should be contextualized as "explanations aligned with annotator reasoning patterns" rather than ground truth. This characteristic applies broadly to supervised explanation methods and highlights the importance of diverse, representative annotation processes. Additionally, extractive rationales capture explicitly highlighted tokens but cannot represent implicit reasoning or world knowledge, a fundamental limitation of token-level annotation schemes rather than the learning approach.

While we provide comprehensive quantitative evaluation (F1, AUROC, precision, recall) and computational benchmarks, complementing these metrics with human-subject studies would strengthen interpretability claims. Such studies could assess whether ExpNet's explanations improve user understanding, decision-making, or trust calibration beyond annotation alignment. This represents valuable future work to validate practical utility in deployed systems.

These considerations frame productive research directions—extending domain coverage, scaling to diverse architectures, enriching annotation practices, and validating with end users—that build upon ExpNet's demonstrated effectiveness at learning generalizable explanation patterns from transformer attention.

\bibliography{custom}

\begin{thebibliography}{36}
\providecommand{\natexlab}[1]{#1}

\bibitem[{Abnar and Zuidema(2020{\natexlab{a}})}]{Abnar2020QuantifyingAF}
Samira Abnar and Willem Zuidema. 2020{\natexlab{a}}.
\newblock \href {https://api.semanticscholar.org/CorpusID:218487351} {Quantifying attention flow in transformers}.
\newblock In \emph{Annual Meeting of the Association for Computational Linguistics}.

\bibitem[{Abnar and Zuidema(2020{\natexlab{b}})}]{attrollout}
Samira Abnar and Willem Zuidema. 2020{\natexlab{b}}.
\newblock \href {https://api.semanticscholar.org/CorpusID:218487351} {Quantifying attention flow in transformers}.
\newblock In \emph{Annual Meeting of the Association for Computational Linguistics}.

\bibitem[{Attanasio et~al.(2022)Attanasio, Pastor, Bonaventura, and Nozza}]{Attanasio2022ferretAF}
Giuseppe Attanasio, Eliana Pastor, Chiara~Di Bonaventura, and Debora Nozza. 2022.
\newblock \href {https://api.semanticscholar.org/CorpusID:251253381} {ferret: a framework for benchmarking explainers on transformers}.
\newblock In \emph{Conference of the European Chapter of the Association for Computational Linguistics}.

\bibitem[{Bach et~al.(2015)Bach, Binder, Montavon, Klauschen, M{\"u}ller, and Samek}]{Bach2015OnPE}
Sebastian Bach, Alexander Binder, Gr{\'e}goire Montavon, Frederick Klauschen, Klaus-Robert M{\"u}ller, and Wojciech Samek. 2015.
\newblock \href {https://api.semanticscholar.org/CorpusID:9327892} {On pixel-wise explanations for non-linear classifier decisions by layer-wise relevance propagation}.
\newblock \emph{PLoS ONE}, 10.

\bibitem[{Bahdanau et~al.(2014)Bahdanau, Cho, and Bengio}]{Bahdanau2014NeuralMT}
Dzmitry Bahdanau, Kyunghyun Cho, and Yoshua Bengio. 2014.
\newblock \href {https://api.semanticscholar.org/CorpusID:11212020} {Neural machine translation by jointly learning to align and translate}.
\newblock \emph{CoRR}, abs/1409.0473.

\bibitem[{Barkan et~al.(2021)Barkan, Hauon, Caciularu, Katz, Malkiel, Armstrong, and Koenigstein}]{Barkan2021GradSAMET}
Oren Barkan, Edan Hauon, Avi Caciularu, Ori Katz, Itzik Malkiel, Omri Armstrong, and Noam Koenigstein. 2021.
\newblock \href {https://api.semanticscholar.org/CorpusID:240230832} {Grad-sam: Explaining transformers via gradient self-attention maps}.
\newblock \emph{Proceedings of the 30th ACM International Conference on Information \& Knowledge Management}.

\bibitem[{Brunner et~al.(2019)Brunner, Liu, Pascual, Richter, Ciaramita, and Wattenhofer}]{Brunner2019OnII}
Gino Brunner, Yang Liu, Damian Pascual, Oliver Richter, Massimiliano Ciaramita, and Roger Wattenhofer. 2019.
\newblock \href {https://api.semanticscholar.org/CorpusID:208100714} {On identifiability in transformers}.
\newblock \emph{arXiv: Computation and Language}.

\bibitem[{Chefer et~al.(2020)Chefer, Gur, and Wolf}]{Transformer‑LRP}
Hila Chefer, Shir Gur, and Lior Wolf. 2020.
\newblock \href {https://api.semanticscholar.org/CorpusID:229297908} {Transformer interpretability beyond attention visualization}.
\newblock \emph{2021 IEEE/CVF Conference on Computer Vision and Pattern Recognition (CVPR)}, pages 782--791.

\bibitem[{Chefer et~al.(2021)Chefer, Gur, and Wolf}]{Chefer2021GenericAE}
Hila Chefer, Shir Gur, and Lior Wolf. 2021.
\newblock \href {https://api.semanticscholar.org/CorpusID:232417173} {Generic attention-model explainability for interpreting bi-modal and encoder-decoder transformers}.
\newblock \emph{2021 IEEE/CVF International Conference on Computer Vision (ICCV)}, pages 387--396.

\bibitem[{Clark et~al.(2019)Clark, Khandelwal, Levy, and Manning}]{Clark2019WhatDB}
Kevin Clark, Urvashi Khandelwal, Omer Levy, and Christopher~D. Manning. 2019.
\newblock \href {https://api.semanticscholar.org/CorpusID:184486746} {What does bert look at? an analysis of bert’s attention}.
\newblock In \emph{BlackboxNLP@ACL}.

\bibitem[{Devlin et~al.(2019)Devlin, Chang, Lee, and Toutanova}]{Devlin2019BERTPO}
Jacob Devlin, Ming-Wei Chang, Kenton Lee, and Kristina Toutanova. 2019.
\newblock \href {https://api.semanticscholar.org/CorpusID:52967399} {Bert: Pre-training of deep bidirectional transformers for language understanding}.
\newblock In \emph{North American Chapter of the Association for Computational Linguistics}.

\bibitem[{DeYoung et~al.(2019)DeYoung, Jain, Rajani, Lehman, Xiong, Socher, and Wallace}]{DeYoung2019ERASERAB}
Jay DeYoung, Sarthak Jain, Nazneen Rajani, Eric~P. Lehman, Caiming Xiong, Richard Socher, and Byron~C. Wallace. 2019.
\newblock \href {https://api.semanticscholar.org/CorpusID:207847663} {Eraser: A benchmark to evaluate rationalized nlp models}.
\newblock In \emph{Annual Meeting of the Association for Computational Linguistics}.

\bibitem[{Doshi-Velez and Kim(2017)}]{DoshiVelez2017TowardsAR}
Finale Doshi-Velez and Been Kim. 2017.
\newblock \href {https://api.semanticscholar.org/CorpusID:11319376} {Towards a rigorous science of interpretable machine learning}.
\newblock \emph{arXiv: Machine Learning}.

\bibitem[{Feng et~al.(2018)Feng, Wallace, II, Iyyer, Rodriguez, and Boyd-Graber}]{Feng2018PathologiesON}
Shi Feng, Eric Wallace, Alvin~Grissom II, Mohit Iyyer, Pedro Rodriguez, and Jordan~Lee Boyd-Graber. 2018.
\newblock \href {https://api.semanticscholar.org/CorpusID:52003282} {Pathologies of neural models make interpretations difficult}.
\newblock In \emph{Conference on Empirical Methods in Natural Language Processing}.

\bibitem[{Herrewijnen et~al.(2024)Herrewijnen, Nguyen, Bex, and van Deemter}]{Herrewijnen2024HumanannotatedRA}
E.~Herrewijnen, Dong Nguyen, Floris Bex, and Kees van Deemter. 2024.
\newblock \href {https://api.semanticscholar.org/CorpusID:270037110} {Human-annotated rationales and explainable text classification: a survey}.
\newblock \emph{Frontiers in Artificial Intelligence}, 7.

\bibitem[{Hooker et~al.(2018)Hooker, Erhan, Kindermans, and Kim}]{Hooker2018ABF}
Sara Hooker, D.~Erhan, Pieter-Jan Kindermans, and Been Kim. 2018.
\newblock \href {https://api.semanticscholar.org/CorpusID:202782699} {A benchmark for interpretability methods in deep neural networks}.
\newblock In \emph{Neural Information Processing Systems}.

\bibitem[{Jacovi and Goldberg(2020)}]{Jacovi2020TowardsFI}
Alon Jacovi and Yoav Goldberg. 2020.
\newblock \href {https://api.semanticscholar.org/CorpusID:215416110} {Towards faithfully interpretable nlp systems: How should we define and evaluate faithfulness?}
\newblock In \emph{Annual Meeting of the Association for Computational Linguistics}.

\bibitem[{Jain and Wallace(2019)}]{Jain2019AttentionIN}
Sarthak Jain and Byron~C. Wallace. 2019.
\newblock \href {https://api.semanticscholar.org/CorpusID:67855860} {Attention is not explanation}.
\newblock In \emph{North American Chapter of the Association for Computational Linguistics}.

\bibitem[{Krippendorff(2011)}]{Krippendorff2011ComputingKA}
Klaus Krippendorff. 2011.
\newblock \href {https://api.semanticscholar.org/CorpusID:59901023} {Computing krippendorff's alpha-reliability}.
\newblock In \emph{Conference}.

\bibitem[{Lin et~al.(2017)Lin, Goyal, Girshick, He, and Doll{\'a}r}]{Lin2017FocalLF}
Tsung-Yi Lin, Priya Goyal, Ross~B. Girshick, Kaiming He, and Piotr Doll{\'a}r. 2017.
\newblock \href {https://api.semanticscholar.org/CorpusID:47252984} {Focal loss for dense object detection}.
\newblock \emph{2017 IEEE International Conference on Computer Vision (ICCV)}, pages 2999--3007.

\bibitem[{Lundberg and Lee(2017)}]{Lundberg2017AUA}
Scott~M. Lundberg and Su-In Lee. 2017.
\newblock \href {https://api.semanticscholar.org/CorpusID:21889700} {A unified approach to interpreting model predictions}.
\newblock In \emph{Neural Information Processing Systems}.

\bibitem[{Mathew et~al.(2020)Mathew, Saha, Yimam, Biemann, Goyal, and Mukherjee}]{Mathew2020HateXplainAB}
Binny Mathew, Punyajoy Saha, Seid~Muhie Yimam, Chris Biemann, Pawan Goyal, and Animesh Mukherjee. 2020.
\newblock \href {https://api.semanticscholar.org/CorpusID:229332119} {Hatexplain: A benchmark dataset for explainable hate speech detection}.
\newblock In \emph{AAAI Conference on Artificial Intelligence}.

\bibitem[{Qiang et~al.(2022)Qiang, Pan, Li, Li, Jang, and Zhu}]{Qiang2022AttCATET}
Yao Qiang, Deng Pan, Chengyin Li, X.~Li, Rhongho Jang, and D.~Zhu. 2022.
\newblock \href {https://api.semanticscholar.org/CorpusID:253115122} {Attcat: Explaining transformers via attentive class activation tokens}.
\newblock In \emph{Neural Information Processing Systems}.

\bibitem[{Ribeiro et~al.(2016)Ribeiro, Singh, and Guestrin}]{Ribeiro2016WhySI}
Marco~Tulio Ribeiro, Sameer Singh, and Carlos Guestrin. 2016.
\newblock \href {https://api.semanticscholar.org/CorpusID:13029170} {“why should i trust you?”: Explaining the predictions of any classifier}.
\newblock \emph{Proceedings of the 22nd ACM SIGKDD International Conference on Knowledge Discovery and Data Mining}.

\bibitem[{Selvaraju et~al.(2016)Selvaraju, Das, Vedantam, Cogswell, Parikh, and Batra}]{Selvaraju2016GradCAMVE}
Ramprasaath~R. Selvaraju, Abhishek Das, Ramakrishna Vedantam, Michael Cogswell, Devi Parikh, and Dhruv Batra. 2016.
\newblock \href {https://api.semanticscholar.org/CorpusID:15019293} {Grad-cam: Visual explanations from deep networks via gradient-based localization}.
\newblock \emph{International Journal of Computer Vision}, 128:336 -- 359.

\bibitem[{Serrano and Smith(2019)}]{Serrano2019IsAI}
Sofia Serrano and Noah~A. Smith. 2019.
\newblock \href {https://api.semanticscholar.org/CorpusID:182953113} {Is attention interpretable?}
\newblock In \emph{Annual Meeting of the Association for Computational Linguistics}.

\bibitem[{Socher et~al.(2013)Socher, Perelygin, Wu, Chuang, Manning, Ng, and Potts}]{Socher2013RecursiveDM}
Richard Socher, Alex Perelygin, Jean Wu, Jason Chuang, Christopher~D. Manning, A.~Ng, and Christopher Potts. 2013.
\newblock \href {https://api.semanticscholar.org/CorpusID:990233} {Recursive deep models for semantic compositionality over a sentiment treebank}.
\newblock In \emph{Conference on Empirical Methods in Natural Language Processing}.

\bibitem[{Song et~al.(2024)Song, Cui, Luo, L{\'e}cu{\'e}, and Li}]{Mask-LRP}
Linxin Song, Yan Cui, Ao~Luo, Freddy L{\'e}cu{\'e}, and Irene Li. 2024.
\newblock \href {https://api.semanticscholar.org/CorpusID:267035104} {Better explain transformers by illuminating important information}.
\newblock In \emph{Findings}.

\bibitem[{Sundararajan et~al.(2017{\natexlab{a}})Sundararajan, Taly, and Yan}]{IntegratedGradients}
Mukund Sundararajan, Ankur Taly, and Qiqi Yan. 2017{\natexlab{a}}.
\newblock \href {https://api.semanticscholar.org/CorpusID:16747630} {Axiomatic attribution for deep networks}.
\newblock In \emph{International Conference on Machine Learning}.

\bibitem[{Sundararajan et~al.(2017{\natexlab{b}})Sundararajan, Taly, and Yan}]{Sundararajan2017AxiomaticAF}
Mukund Sundararajan, Ankur Taly, and Qiqi Yan. 2017{\natexlab{b}}.
\newblock \href {https://api.semanticscholar.org/CorpusID:16747630} {Axiomatic attribution for deep networks}.
\newblock In \emph{International Conference on Machine Learning}.

\bibitem[{Vaswani et~al.(2017)Vaswani, Shazeer, Parmar, Uszkoreit, Jones, Gomez, Kaiser, and Polosukhin}]{Vaswani2017AttentionIA}
Ashish Vaswani, Noam~M. Shazeer, Niki Parmar, Jakob Uszkoreit, Llion Jones, Aidan~N. Gomez, Lukasz Kaiser, and Illia Polosukhin. 2017.
\newblock \href {https://api.semanticscholar.org/CorpusID:13756489} {Attention is all you need}.
\newblock In \emph{Neural Information Processing Systems}.

\bibitem[{Vig(2019)}]{Vig2019AMV}
Jesse Vig. 2019.
\newblock \href {https://api.semanticscholar.org/CorpusID:189762556} {A multiscale visualization of attention in the transformer model}.
\newblock \emph{ArXiv}, abs/1906.05714.

\bibitem[{Voita et~al.(2019)Voita, Talbot, Moiseev, Sennrich, and Titov}]{Voita2019AnalyzingMS}
Elena Voita, David Talbot, F{\'e}dor Moiseev, Rico Sennrich, and Ivan Titov. 2019.
\newblock \href {https://api.semanticscholar.org/CorpusID:162183964} {Analyzing multi-head self-attention: Specialized heads do the heavy lifting, the rest can be pruned}.
\newblock \emph{ArXiv}, abs/1905.09418.

\bibitem[{Warstadt et~al.(2018)Warstadt, Singh, and Bowman}]{warstadt2018neural}
Alex Warstadt, Amanpreet Singh, and Samuel~R Bowman. 2018.
\newblock Neural network acceptability judgments.
\newblock \emph{arXiv preprint arXiv:1805.12471}.

\bibitem[{Wiegreffe and Pinter(2019)}]{Wiegreffe2019AttentionIN}
Sarah Wiegreffe and Yuval Pinter. 2019.
\newblock \href {https://api.semanticscholar.org/CorpusID:199552244} {Attention is not not explanation}.
\newblock In \emph{Conference on Empirical Methods in Natural Language Processing}.

\bibitem[{Xu et~al.(2015)Xu, Ba, Kiros, Cho, Courville, Salakhutdinov, Zemel, and Bengio}]{Xu2015ShowAA}
Ke~Xu, Jimmy Ba, Ryan Kiros, Kyunghyun Cho, Aaron~C. Courville, Ruslan Salakhutdinov, Richard~S. Zemel, and Yoshua Bengio. 2015.
\newblock \href {https://api.semanticscholar.org/CorpusID:1055111} {Show, attend and tell: Neural image caption generation with visual attention}.
\newblock In \emph{International Conference on Machine Learning}.

\end{thebibliography}

\newpage
\appendix

\section{Annotation Methodology and Reliability}
\label{app:annotation}

\subsection{LLM-Based Annotation Procedure}

For SST-2 and CoLA, which lack ground-truth rationales, we employed a systematic three-stage annotation procedure using large language models:

\paragraph{Stage 1: Calibration on Human-Annotated Data} 
To ground the procedure in verifiable signal, we calibrated on HateXplain, which provides human token/phrase-level rationales. We evaluated multiple LLMs—GPT-4o, GPT-3.5-turbo, Claude-3-5-Haiku-20241022, and Claude-3-5-Sonnet-20241022—under several prompt variants to predict labels and highlight supporting tokens. We quantitatively compared token-level precision, recall, and F1 against the human spans. Based on this calibration, we identified Claude-3-5-Sonnet-20241022 as the most reliable annotator and established best practices for prompt engineering.

\paragraph{Stage 2: Prompt Design and Validation} 
For SST-2 and CoLA, we curated small, manually labeled seed sets with human-annotated rationale examples (30 examples for SST-2, evenly split across classes; 30 examples for CoLA focusing on unacceptable sentences) to anchor the notion of useful token-level rationales. Guided by these examples, we iteratively refined task-specific prompts instructing the LLM to (a) predict the task label, (b) highlight tokens that justify that prediction, and (c) provide a brief explanation for their reasoning. 

We developed seven prompt variants and selected the best-performing version based on the validation sets. For SST-2, the selected prompt achieved micro-F1 = 0.66 and macro-F1 = 0.69 (precision = 0.58, recall = 0.77) on identifying important tokens. For CoLA, we achieved micro-F1 = 0.88 (precision = 0.85, recall = 0.92). These figures reflect expected task differences: sentiment cues are often short and context-dependent (harder to capture consistently), whereas grammatical violations tend to be localized and more systematically recoverable.

\paragraph{Stage 3: Large-Scale Annotation with Quality Controls} 
Using the selected prompts, we annotated the training splits at scale. To avoid propagating rationales for incorrect task predictions, we retained only instances where the LLM's predicted label matched the gold label (label-agreement filter). We then applied light spot-checking to remove outputs with incoherent or diffuse spans. For SST-2, we limited LLM annotation to approximately 6,000 train examples due to API cost constraints. For CoLA, we annotated 2,426 training instances (primarily unacceptable examples). The final annotation counts are: SST-2 (5,971 train / 395 validation), CoLA (2,426 train / 101 validation), and HateXplain (8,042 train / 734 validation), with test sets created by splitting original validation data to ensure label balance.

\begin{table}[ht]
\centering
\footnotesize
\caption{LLM annotation calibration and selection results.}
\label{tab:llm_calibration}
\begin{tabular}{@{}llccc@{}}
\toprule
\textbf{Dataset} & \textbf{LLM} & \textbf{F1} & \textbf{Seeds} & \textbf{Final} \\
\midrule
\multirow{3}{*}{SST-2} & claude-3-5-sonnet & \textbf{0.69} & \multirow{3}{*}{30} & \multirow{3}{*}{5,971/395} \\
                        & claude-3-5-haiku  & 0.68 & & \\
                        & gpt-3.5-turbo     & 0.60 & & \\
\midrule
\multirow{3}{*}{CoLA}  & claude-3-5-sonnet & \textbf{0.81} & \multirow{3}{*}{30} & \multirow{3}{*}{2,426/101} \\
                        & claude-3-5-haiku  & 0.63 & & \\
                        & gpt-3.5-turbo     & 0.37 & & \\
\midrule
HateXplain & Human & --- & --- & 8,042/734 \\
\bottomrule
\end{tabular}
\end{table}

\subsection{Inter-Annotator Agreement Analysis}

To assess the reliability and consistency of our rationale annotations, we computed Krippendorff's Alpha \cite{Krippendorff2011ComputingKA}, a robust inter-annotator agreement coefficient that handles multiple annotators and accounts for chance agreement. For SST-2 and CoLA, we had two human annotators independently verify a subset of LLM-generated rationales on the test sets (195 examples for SST-2, 51 for CoLA). For HateXplain, we report the published inter-annotator agreement from the original dataset \cite{Mathew2020HateXplainAB}, which involved three annotators per example.

\begin{table}[ht]
\centering
\footnotesize
\caption{Inter-annotator agreement and average rationale length (K).}
\label{tab:interannotator_full}
\begin{tabular}{@{}lccc@{}}
\toprule
\textbf{Dataset} & \textbf{Test Samples} & \textbf{$\alpha$ (Agreement)} & \textbf{Avg. K} \\
\midrule
SST-2       & 195 & 0.84 (substantial) & 3 \\
CoLA        & 51  & 0.64 (moderate)    & 15 \\
HateXplain  & 726 & 0.46 (moderate)    & 9 \\
\bottomrule
\end{tabular}
\end{table}

The higher agreement for SST-2 reflects clearer sentiment-bearing tokens (e.g., ``excellent'', ``terrible''), while CoLA's moderate agreement stems from linguistic nuance in grammatical judgments. HateXplain's lower agreement is consistent with the subjective nature of hate speech annotation, where annotators may focus on different contextual cues. These agreement levels are comparable to or exceed typical values reported in rationale annotation studies \cite{DeYoung2019ERASERAB}, supporting the reliability of our evaluation data.

\subsection{Baseline Binarization and Alignment Procedures}
\label{app:baselines}

\subsection{Binarization of Continuous Scores}

To enable fair comparison with binary human rationales, we binarize the continuous attribution scores produced by each baseline using the top-K selection heuristic from \citet{Attanasio2022ferretAF}. For each dataset, we first compute the average number of annotated important words ($K$) in the training split. We then sort the tokens of each test instance by their attribution scores (descending) and label the top-$K$ tokens as important (1), assigning 0 to the remainder. Thus, each explainer highlights, on average, the same count of tokens that humans deemed relevant in that dataset. The empirically derived $K$ values are: $K_{\text{SST-2}} = 3$, $K_{\text{CoLA}} = 15$, and $K_{\text{HateXplain}} = 9$. This ranking-based thresholding is applied independently for every baseline and dataset, yielding comparable binary rationales across all methods.

\textbf{Procedure per instance for baseline methods:}
\begin{enumerate}
    \item Rank tokens by attribution score $s_k$ (descending)
    \item Exclude negative scores if a baseline method produces them
    \item Select the top-K tokens as $\hat{R}$
    \item Tie-breaking: if several tokens share the K-th score, break ties by (i) earlier position in the sequence, then (ii) token id to ensure determinism
    \item Short sequences: if $|\mathcal{T}| < K$, set $\hat{R} = \mathcal{T}$
\end{enumerate}

\subsection{Token-to-Word Alignment}

All baseline methods operate over BERT's WordPiece sub-tokens \cite{Devlin2019BERTPO}. For supervision and evaluation with word-level references, we employ standard alignment procedures: \textbf{Label projection (training):} If any sub-token of a word is labeled positive, all its sub-tokens inherit that label. \textbf{Score aggregation (evaluation):} A word's score is the maximum over its sub-token scores; binary predictions $\hat{R}$ from baselines are assessed at the word level. This alignment ensures baseline predictions are evaluated against the same granularity as human references and that evaluation is invariant to sub-tokenization differences across baseline methods.

\section{Evaluation Metric Definitions}
\label{app:metrics}

\subsection{Precision, Recall, and F1-Score}

For a test instance with token set $\mathcal{T}$ and reference rationale $R \subseteq \mathcal{T}$, each method produces a binary prediction $\hat{R} \subseteq \mathcal{T}$. With true positives $TP = |\hat{R} \cap R|$, false positives $FP = |\hat{R} \setminus R|$, and false negatives $FN = |R \setminus \hat{R}|$, we compute:
\begin{align}
P &= \frac{TP}{TP + FP}, \quad R = \frac{TP}{TP + FN} \\
F_1 &= \frac{2PR}{P + R} = \frac{2TP}{2TP + FP + FN}
\end{align}

We report dataset-level F1 by summing $TP$, $FP$, $FN$ over all evaluated instances before computing $F_1$, following common practice in rationale evaluation \cite{DeYoung2019ERASERAB}.

\subsection{AUROC (Area Under the ROC Curve)}

AUROC is a threshold-independent metric that assesses the ranking quality of continuous importance scores across all possible thresholds. Treating each token as positive if $k \in R$, we vary a threshold $\tau$ on scores $s_k$ to obtain $\hat{R}(\tau) = \{k : s_k \geq \tau\}$ and compute the true positive rate (TPR) and false positive rate (FPR):
\begin{align}
\text{TPR}(\tau) &= \frac{|\hat{R}(\tau) \cap R|}{|R|} \\
\text{FPR}(\tau) &= \frac{|\hat{R}(\tau) \setminus R|}{|\mathcal{T} \setminus R|}
\end{align}

AUROC is the area under the ROC curve formed by plotting TPR against FPR as $\tau$ varies. For dataset-level AUROC, we pool all token scores and labels from all evaluated instances. AUROC complements F1 by capturing how well an explainer orders true rationale tokens ahead of non-rationale tokens, independent of any particular threshold choice. A higher AUROC indicates that the method consistently assigns higher scores to important tokens than to unimportant ones.

\section{Performance metrics}
\label{sec:performance_metrics}

\begin{figure*}[t]
  \includegraphics[width=0.32\linewidth]{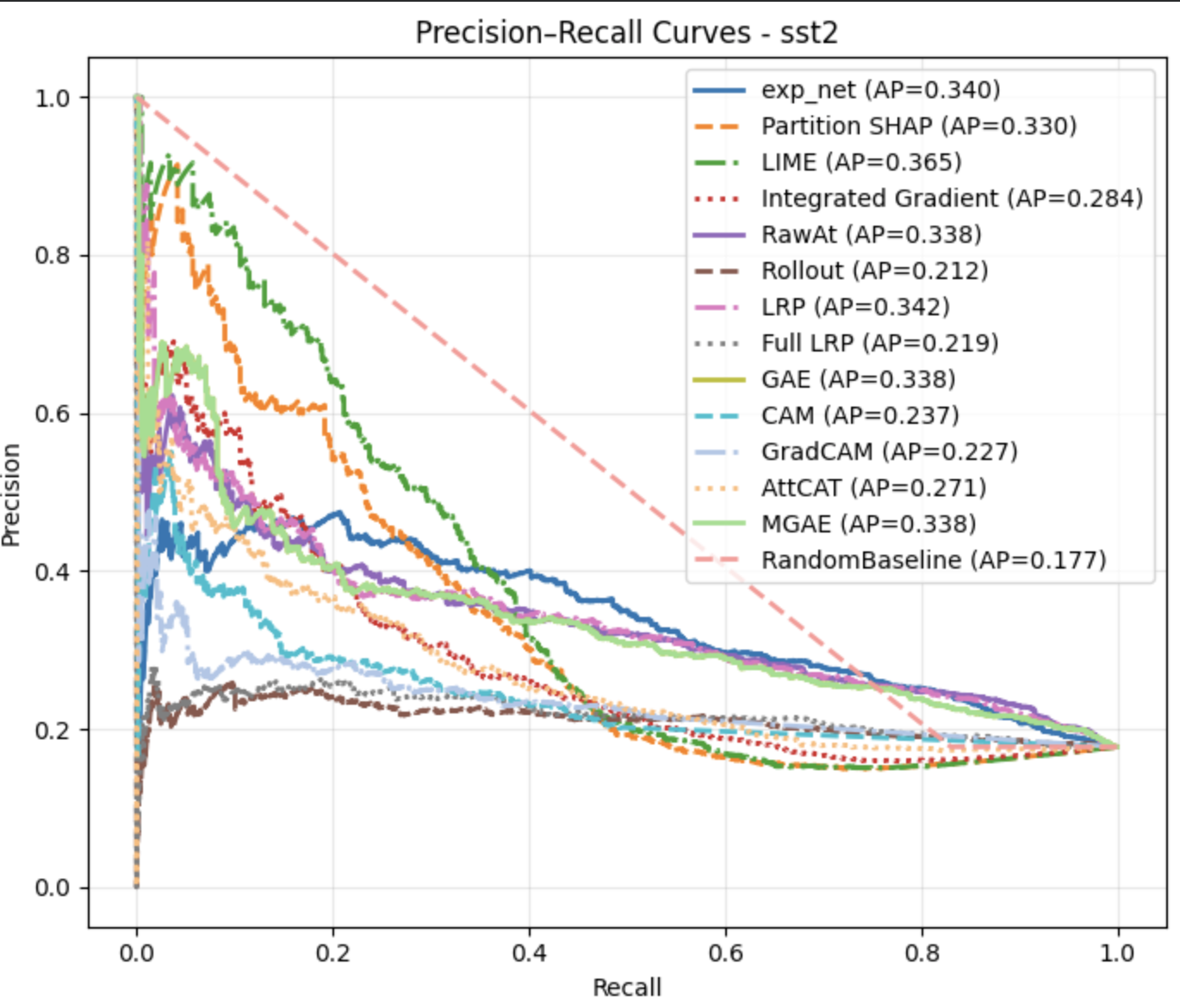} \hfill
  \includegraphics[width=0.32\linewidth]{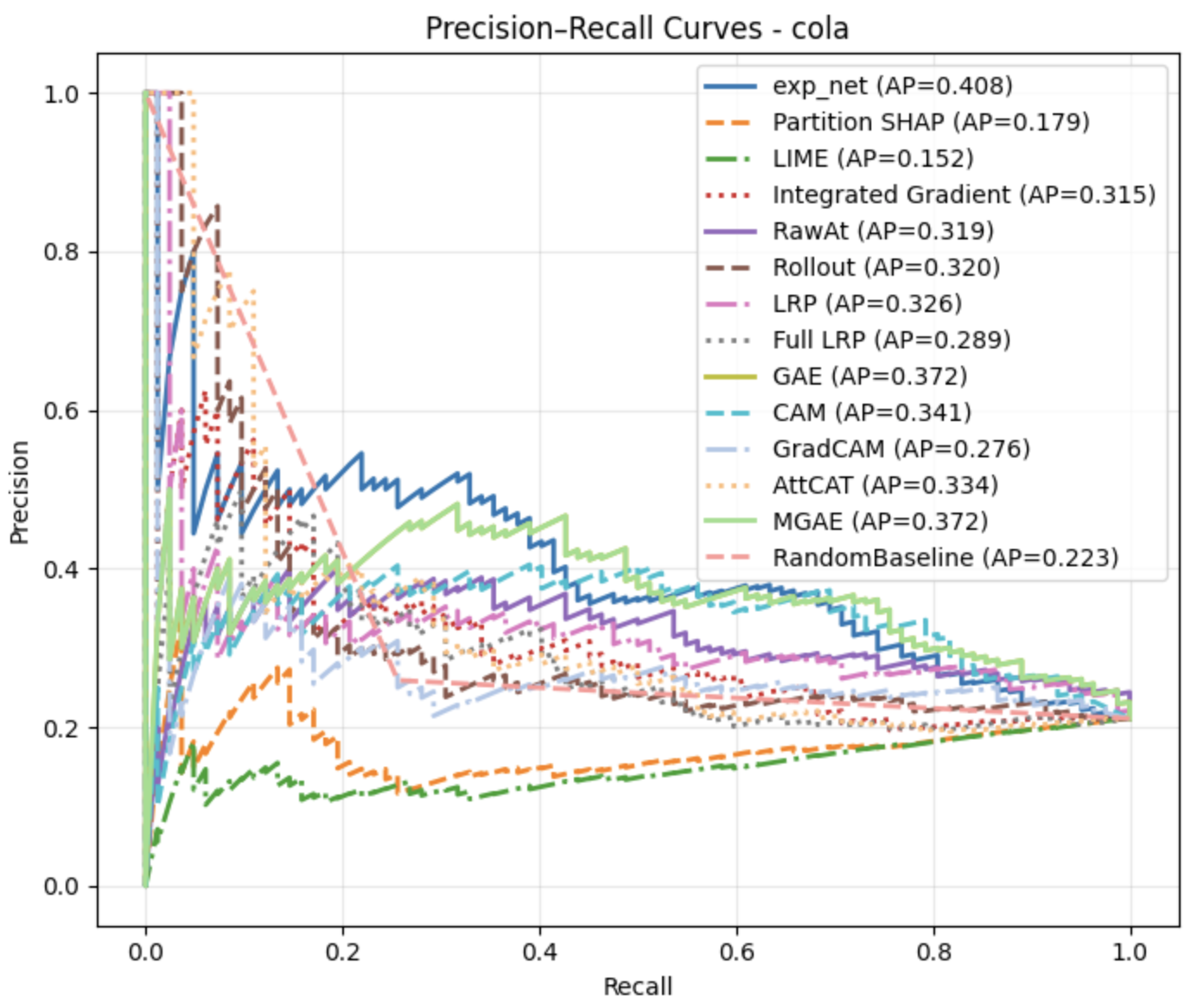} \hfill
  \includegraphics[width=0.32\linewidth]{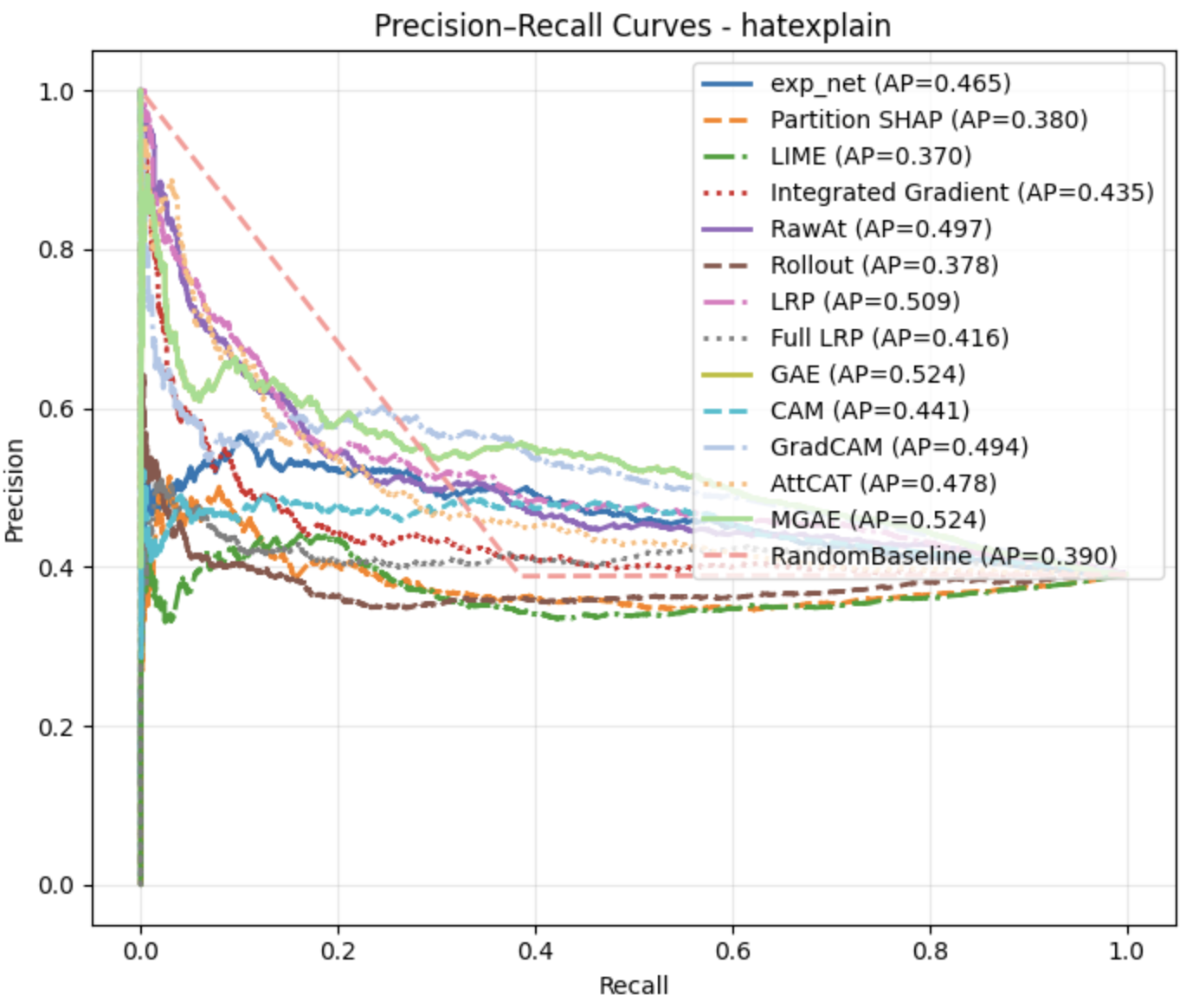}
  \caption{\label{aupr}
  AUPR values across datasets}
\end{figure*}

Performance metrics (F1-score, Precision, Recall) with confidence intervals for ExpNet compared to baseline methods under cross-dataset evaluation presented in Table \ref{tab:model_performance_full}. AUROC values across dataset presented in Figure \ref{aupr}.

\begin{table*}[t]
  \centering
  \small
  \begin{tabular}{l ccc ccc ccc}
    \toprule
    & \multicolumn{3}{c}{\textbf{SST-2}} & \multicolumn{3}{c}{\textbf{CoLA}} & \multicolumn{3}{c}{\textbf{HateXplain}} \\
    \cmidrule(lr){2-4} \cmidrule(lr){5-7} \cmidrule(lr){8-10}
    \textbf{Explainer} & \textbf{F1} & \textbf{P} & \textbf{R} & \textbf{F1} & \textbf{P} & \textbf{R} & \textbf{F1} & \textbf{P} & \textbf{R} \\
    \midrule
    RandomBaseline & 0.258 & 0.387 & 0.293 & 0.259 & 0.388 & 0.178 & 0.256 & 0.385 & $\mathbf{0.830}$ \\
    SHAP & $0.330 \pm 0.033$ & $0.392$ & $0.284$ & $0.330 \pm 0.056$ & $0.202$ & $0.902$ & $0.276 \pm 0.007$ & $0.384$ & $0.215$ \\
    LIME & $0.347 \pm 0.033$ & $\mathbf{0.408}$ & $0.301$ & $0.323 \pm 0.053$ & $0.198$ & $0.878$ & $0.290 \pm 0.007$ & $0.376$ & $0.236$ \\
    Integrated Gradient & $0.287 \pm 0.032$ & $0.332$ & $0.254$ & $0.342 \pm 0.054$ & $0.209$ & $0.939$ & $0.345 \pm 0.008$ & $0.433$ & $0.286$ \\
    RawAt & $0.327 \pm 0.029$ & $0.364$ & $0.295$ & $0.353 \pm 0.054$ & $0.214$ & $\mathbf{1.000}$ & $0.362 \pm 0.007$ & $0.541$ & $0.271$ \\
    Rollout & $0.133 \pm 0.025$ & $0.150$ & $0.119$ & $0.347 \pm 0.055$ & $0.210$ & $0.987$ & $0.356 \pm 0.007$ & $0.446$ & $0.297$ \\
    LRP & $0.339 \pm 0.030$ & $0.377$ & $0.307$ & $0.355 \pm 0.054$ & $0.216$ & $\mathbf{1.000}$ & $0.372 \pm 0.008$ & $\mathbf{0.544}$ & $0.282$ \\
    Full LRP & $0.218 \pm 0.030$ & $0.245$ & $0.196$ & $0.336 \pm 0.055$ & $0.204$ & $0.926$ & $0.346 \pm 0.007$ & $0.407$ & $0.300$ \\
    GAE & $0.350 \pm 0.030$ & $0.391$ & $0.316$ & $0.354 \pm 0.053$ & $0.215$ & $\mathbf{1.000}$ & $0.391 \pm 0.007$ & $0.528$ & $0.310$ \\
    CAM & $0.243 \pm 0.032$ & $0.287$ & $0.211$ & $0.355 \pm 0.054$ & $0.216$ & $\mathbf{1.000}$ & $0.332 \pm 0.007$ & $0.421$ & $0.273$ \\
    GradCAM & $0.234 \pm 0.031$ & $0.271$ & $0.205$ & $0.356 \pm 0.056$ & $0.216$ & $0.987$ & $0.396 \pm 0.008$ & $0.521$ & $0.320$ \\
    AttCAT & $0.280 \pm 0.034$ & $0.360$ & $0.228$ & $0.345 \pm 0.055$ & $0.211$ & $0.963$ & $0.340 \pm 0.007$ & $0.461$ & $0.268$ \\
    MGAE & $0.350 \pm 0.030$ & $0.391$ & $0.316$ & $0.354 \pm 0.053$ & $0.215$ & $\mathbf{1.000}$ & $0.391 \pm 0.007$ & $0.528$ & $0.310$ \\
    \midrule
    ExpNet & $\mathbf{0.398 \pm 0.024}$ & $0.285$ & $\mathbf{0.660}$ & $\mathbf{0.468 \pm 0.079}$ & $\mathbf{0.349}$ & $0.707$ & $\mathbf{0.473 \pm 0.007}$ & $0.466$ & 0.478 \\
    \bottomrule
  \end{tabular}
  \caption{\label{tab:model_performance_full}
    Performance metrics (F1-score, Precision, Recall) with confidence intervals for ExpNet compared to baseline methods under cross-dataset evaluation. ExpNet is trained on two datasets and evaluated on the held-out third dataset. RandomBaseline is reported as a single deterministic run.
  }
\end{table*}

\section{Computational Efficiency Details}
\label{sec:computational_details}

Table~\ref{tab:inference_speed} reports the inference throughput (examples processed per second) for ExpNet and all thirteen baseline explanation methods. All measurements were conducted on the same hardware configuration using identical batch sizes to ensure fair comparison. ExpNet achieves 13.889 examples/second, making it approximately 70× faster than LIME, 12× faster than SHAP, and 5× faster than relevance propagation methods (LRP, GAE, MGAE). The attention-based methods (RawAt, Rollout) achieve the highest throughput due to minimal computation, but at the cost of significantly lower F1 scores as shown in Table~\ref{quantitative-results}. ExpNet strikes an optimal balance between explanation quality and computational efficiency.

\begin{table}[ht]
  \centering
  \begin{tabular}{lc}
    \hline
    \textbf{Method} & \textbf{Throughput [ex/s]} $\uparrow$ \\
    \hline
    SHAP            & 1.067 \\
    LIME            & 0.120 \\
    Integrated Gradient & 0.197 \\
    RawAt           & 19.23 \\
    Rollout         & 15.38 \\
    LRP             & 2.873 \\
    FullLRP         & 2.890 \\
    GAE             & 2.832 \\
    MGAE            & 2.538 \\
    CAM             & 2.923 \\
    GradCAM         & 2.840 \\
    AttCAT          & 7.194 \\
    \hline
    \textbf{ExpNet} & \textbf{13.889} \\
    \hline
  \end{tabular}
  \caption{\label{tab:inference_speed}
    Inference throughput comparison across all explanation methods. Higher values indicate faster processing. ExpNet achieves near-attention-based speed while maintaining substantially higher F1 scores than all baselines.
  }
\end{table}

Model-agnostic methods require the most computation: LIME samples thousands of perturbed inputs and trains local surrogate models for each example, achieving only 0.120 ex/s. SHAP computes Shapley values through combinatorial evaluation, reaching 1.067 ex/s. Integrated Gradient requires multiple forward passes with interpolated inputs (0.197 ex/s). Relevance propagation methods (LRP, FullLRP, GAE, MGAE) perform backward passes with specialized propagation rules, achieving 2.5-2.9 ex/s. Gradient-based methods vary: GradCAM and CAM ($\sim$2.9 ex/s) compute gradients with respect to activations, while AttCAT (7.194 ex/s) uses more efficient gradient approximations. Attention-based methods (RawAt: 19.23 ex/s, Rollout: 15.38 ex/s) simply aggregate pre-computed attention weights with minimal overhead.

ExpNet's 13.889 ex/s throughput approaches attention-based methods' speed while delivering F1 scores 14-31\% higher than the best baselines (Table~\ref{quantitative-results}). This efficiency advantage makes ExpNet practical for real-time applications requiring both speed and accuracy, such as interactive model debugging tools, user-facing explanation interfaces, and large-scale model auditing.

\section{Qualitative Examples}
\label{sec:appendix}

Below are representative examples showing how each explainer attributes importance to tokens across the three evaluation datasets. Gold-standard human annotations are shown in the top row for reference. These examples illustrate the qualitative differences between ExpNet and baseline methods beyond aggregate metrics.

\begin{figure*}[t]
\centering
\includegraphics[width=0.95\textwidth, height=0.35\textheight, keepaspectratio]{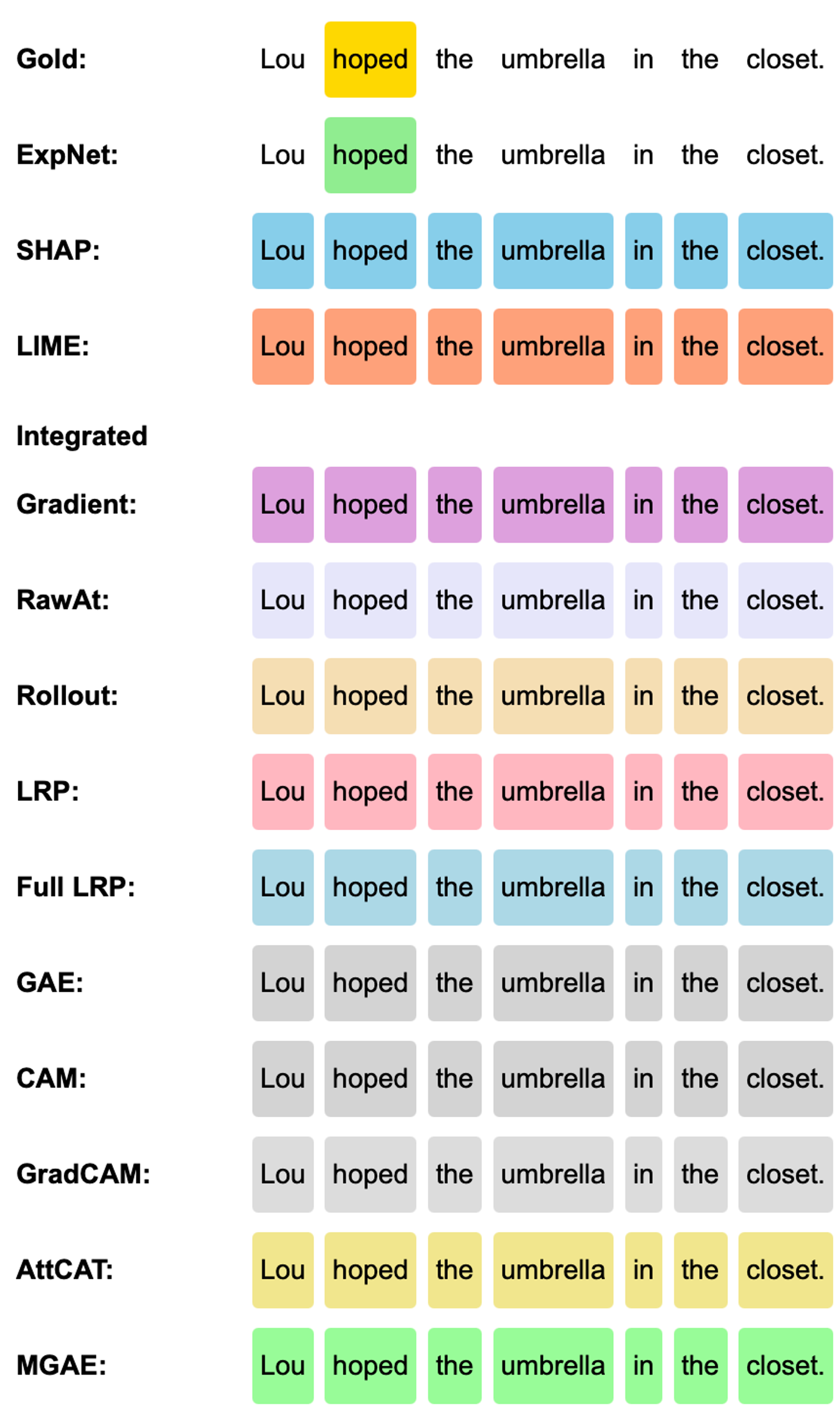}
\caption{Example from CoLA highlighting grammatical acceptability cues. The sentence contains an incomplete verb phrase that determines grammaticality. ExpNet closely matches the gold standard by precisely identifying "hoped" as the critical token. Many baselines either over-highlight the full sentence (LIME, SHAP, Integrated Gradient, LRP, Full LRP, GAE, CAM, GradCAM, AttCAT, MGAE) or fail to identify any specific tokens (RawAt, Rollout), failing to isolate the specific grammatical violation that human annotators identified.}
\label{fig:ex_cola}
\end{figure*}

\begin{figure*}[t]
\centering
\includegraphics[width=0.95\textwidth, height=0.35\textheight, keepaspectratio]{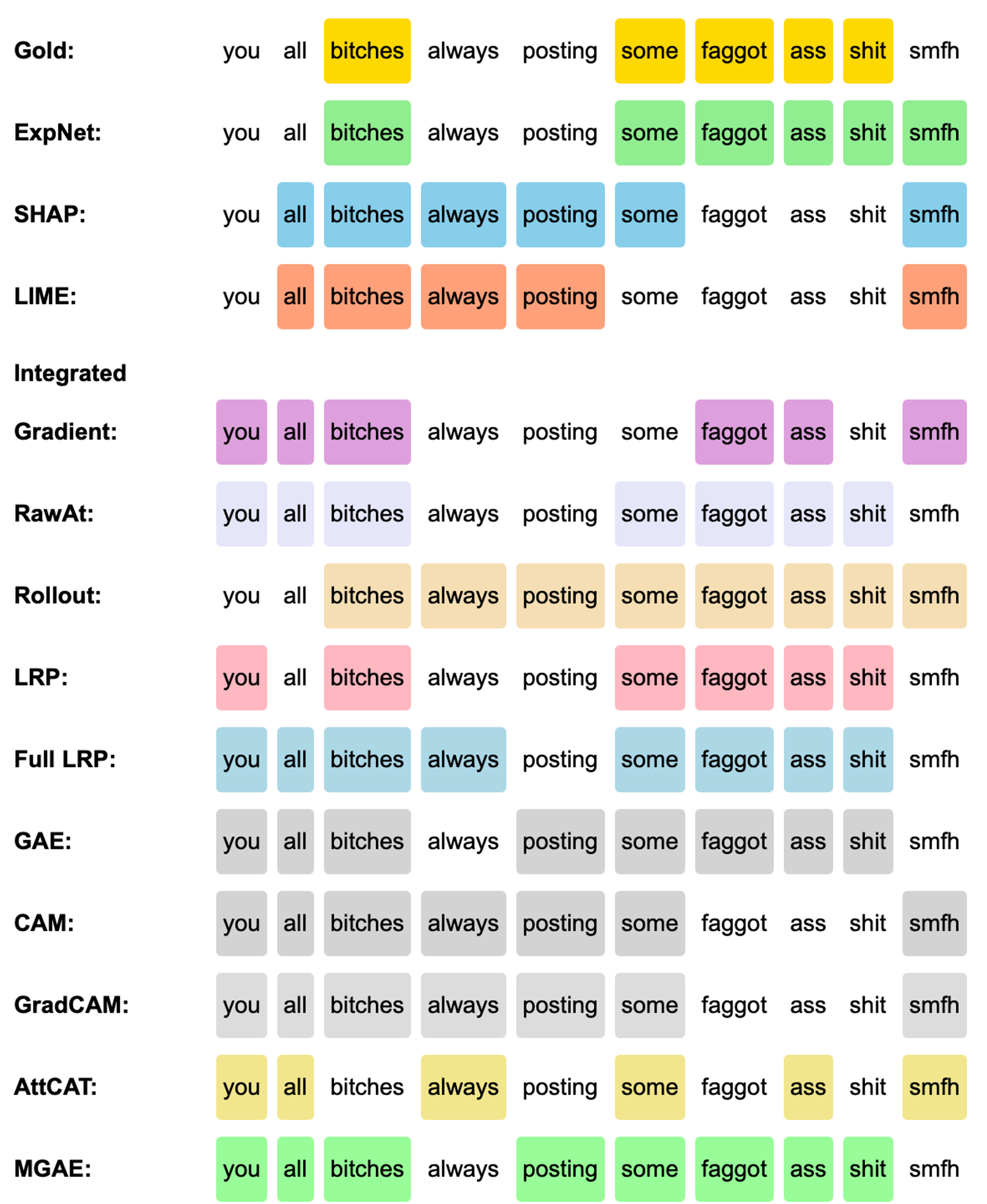}
\caption{Word importance attribution comparisons on a toxic example from the HateXplain dataset. Gold-standard human annotations highlight slurs and hate-inducing terms that constitute the offensive content. ExpNet aligns well with human judgments, correctly identifying the key toxic tokens. Some gradient-based methods (MGAE, LRP, Integrated Gradient) also capture relevant terms, though with varying precision. Model-agnostic methods (SHAP, LIME, Full LRP) highlight irrelevant tokens or produce overly diffuse attributions across most of the sentence, failing to isolate the specific language making the content hateful.}
\label{fig:ex_hatexplain}
\end{figure*}

\begin{figure*}[t]
\centering
\includegraphics[width=0.95\textwidth, height=0.35\textheight, keepaspectratio]{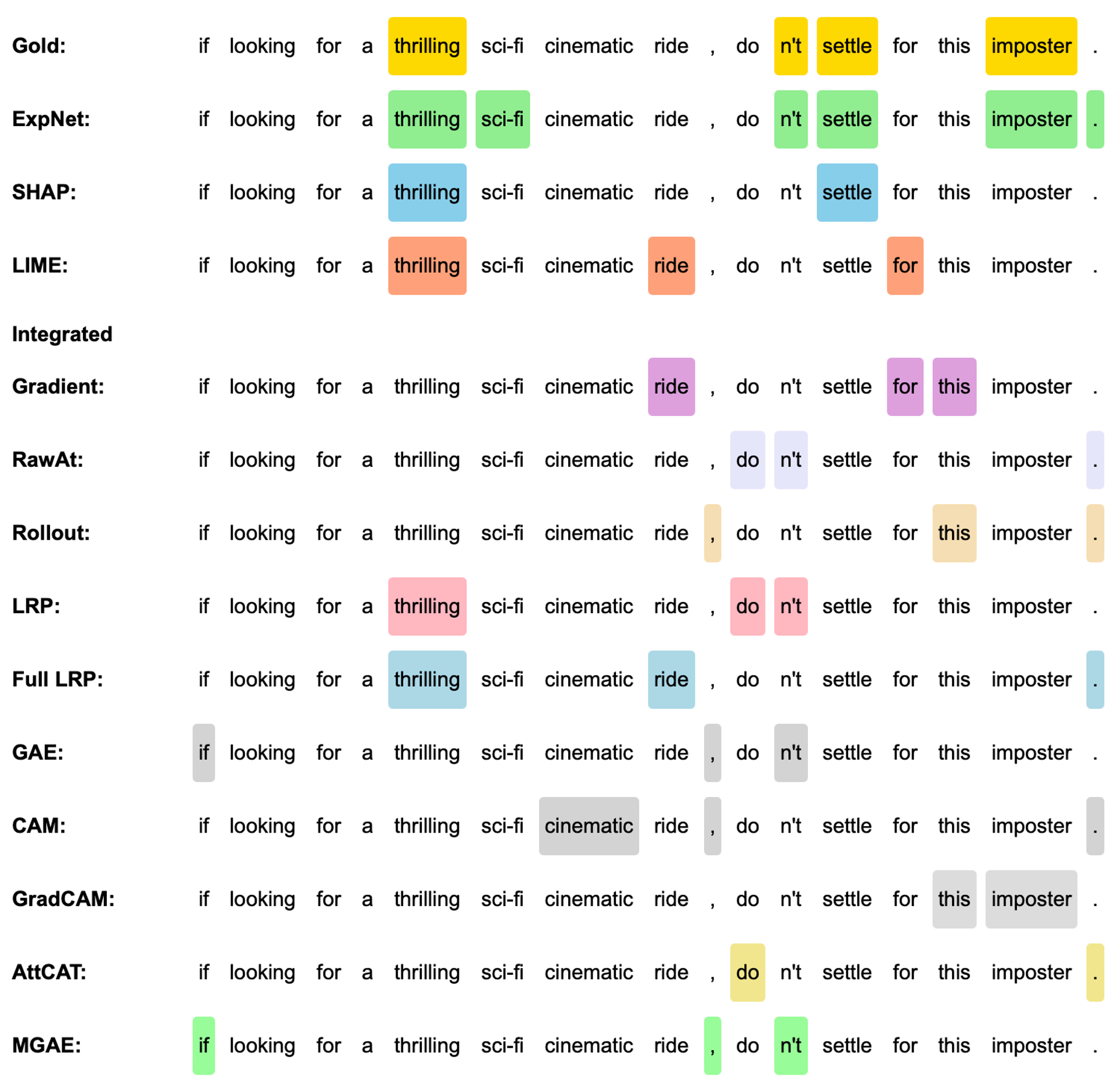}
\caption{Attribution visualization for a movie review from SST-2. The review contains both factual content and sentiment-bearing language. ExpNet highlights semantically loaded tokens such as "thrilling", "sci-fi", and sentiment indicators like "n't", "settle", and "imposter" that drive the negative classification, closely matching human rationales. In contrast, perturbation-based methods like LIME and gradient methods like Integrated Gradient focus on tokens like "ride" and "for" that may be salient to the model but are not the primary sentiment drivers that humans identify. Attention-based methods (RawAt, Rollout) and other gradient methods (GAE, CAM, AttCAT) distribute importance broadly or focus on isolated tokens without discriminating between content and sentiment indicators.}
\label{fig:ex_sst2}
\end{figure*}

These qualitative examples demonstrate patterns consistent with the quantitative results: ExpNet's learned mapping from attention features to importance scores more closely approximates human judgment than fixed heuristics (attention-based), perturbation schemes (model-agnostic), or gradient-based approximations. The examples also reveal task-specific challenges: on CoLA, identifying single grammatical violations requires precision; on HateXplain, distinguishing toxic from non-toxic language requires semantic understanding; on SST-2, separating sentiment from factual content requires nuanced interpretation. ExpNet handles all three patterns more consistently than baselines, supporting the claim that supervised learning on human rationales captures generalizable explanation patterns.

\end{document}